\documentclass[10pt]{article}
\usepackage[a4paper,hmargin=3cm,vmargin=3.0cm,includeheadfoot]{geometry}
\usepackage[utf8]{inputenc}
\usepackage[english]{babel}
\usepackage{float}
\usepackage[font=small]{caption}
\usepackage[affil-it]{authblk}
\usepackage{hyperref}
\usepackage{textgreek}

\usepackage[super]{natbib}
\usepackage{amsmath}
\usepackage{graphicx}
\usepackage[T1]{fontenc}
\usepackage[utf8]{inputenc} % assuming your file is encoded in UTF-8

\title{Attention, not scale, drives human-AI alignment in multimodal language prediction}
\author[1,*]{Viktor Kewenig}
\author[2]{Andrew Lampinen}
\author[3]{Samuel A. Nastase}
\author[1]{Christopher Edwards}
\author[4]{Quitterie D'Elascombe}
\author[1]{Akilles Richardt}
\author[1]{Jeremy I. Skipper}
\author[1]{Gabriella Vigliocco}
\affil[1]{Psychology and Language Science, Experimental Psychology, University College London, London, UK}
\affil[2]{Google Deepmind, Mountain View, US}
\affil[3]{Princeton Neuroscience Institute, Princeton University, Princeton, NJ, USA}
\affil[4]{Computer Science Department, Exeter University}
\affil[*]{Corresponding author: ucjuvnk@ucl.ac.uk}

\begin{document} 
\date{}
\maketitle

\section*{Abstract}
Humans routinely draw on visual context to predict upcoming words. To what extent current vision-language models produce comparable behaviour is unclear. Here we placed five state-of-the-art pretrained systems side-by-side with 600 human participants in a web-based Visual-World Paradigm. On each of 100 six-second movie clips, models and participants received either text only or synchronised video + text and judged how likely a specified target word was to appear next; human eye movements were tracked throughout. Adding visual context increased model–human alignment in predictability ratings across all architectures (average $\Delta r \approx .18$) with no impact of parameter size. When visual context was informative, transformer attention significantly increased alignment. Attention maps from two transformer models corresponded with human gaze, explaining up to 70\% of the inter-participant variance when the scene contained informative cues. Notably, cross-modal attention reliably tracked anticipatory human fixations on semantic cues. These results suggest that current transformer-based vision–language models can approximate human behaviour exploiting visual context during language prediction - and that selective attention to informative cues, not sheer model scale, is the principal driver of this alignment.

\section*{Introduction}
Recent advances in generative artificial intelligence (genAI) have stimulated debate over the extent to which language processing in computational models is comparable to human cognition.\cite{mitchell_debate_2023,mahowald_dissociating_2023,liu_brainclip_2023,kauf_event_2022,zhang_visual_2022,michelmann_large_2023} For example, direct brain-model comparisons suggest that hierarchical predictive processing, driven by contextual linguistic information, underlies language processing in both humans and AI. \cite{schrimpf_neural_2021,goldstein_shared_2022,caucheteux_long-range_2021}. On the other hand, LLMs are found to accurately simulate human behaviour from classic experiments in economic, psycholinguistic, and social psychology experiments \cite{pmlr-v202-aher23a, leng2024llmagentsexhibitsocial, petrov2024limitedabilityllmssimulate, niu2024largelanguagemodelscognitive, Shaki_2023}, such as theory of mind tasks \cite{kosinski2024evaluating}.

One limitation of many of these studies is that they focus on unimodal models (text-based) and unimodal (text- or audio-based) experimental stimuli (but see \citet{dong_vision-language_2023}). Yet in ecological settings, human language input is often accompanied by information from other modalities (e.g. visual) \cite{vigliocco2014language}. Here we assess how additional visual context influences predictive language processing in computational models. We use the term paradigm to denote the input configuration presented to the same pretrained neural network for extracting predictions: a unimodal paradigm supplies text only, whereas a multimodal paradigm supplies synchronised visual information and text. We operationalise different comparisons with human behavioural data using a large-scale online experiment where participants watched 6 s movie clips (or saw only the visual stimulus / listened to an audio-only version) while rating target-word predictability and supplying webcam eye-tracking. 

Theory and empirical evidence suggests that visual information may be an important factor influencing preditive language processing in humans. There is ample behavioural \cite{paivio_imagery_2013, barsalou_perceptual_1999, barsalou_grounded_2008, vigliocco_toward_2009, vigliocco_language_2014, andrews_integrating_2009, andrews_reconciling_2014, louwerse_linguistic_2010, kousta_representation_2011, riordan_redundancy_2011,morita2025genaireading} and neurobiological \cite{kewenig_when_2023, hahamy_human_2023, zadbood_neural_2022, baldassano_representation_2018, ben-yakov_hippocampal_2018} evidence that integrating linguistic input with visual context during language processing places constraints on the space of upcoming semantic content, reducing cognitive load and improving comprehension. 

However, it does not follow that adding visual context to language processing in LLMs equally improves predictions. Integrating multimodal information in computational models is not without challenges; additional modalities can introduce noise that may hinder model performance. Irrelevant or distracting visual information can lead to misalignment between modalities, causing models to focus on inconsequential features.\cite{wang2019words} The complexity of effectively fusing heterogeneous data sources can result in increased computational overhead and potential overfitting, especially when models are not properly regularized.\cite{ramachandram_deep_2017} Moreover, conflicting information between modalities may confuse the model, leading to degraded performance compared to unimodal counterparts.\cite{baltruvsaitis_multimodal_2019}

For these reasons, the capacity to process both visual and linguistic input by itself may provide little benefit for predictive language processing if this information is not integrated appropriately. In humans, paying attention to salient visual cues is an effective mechanism of multimodal integration during language processing, narrowing down the space of upcoming semantic content. Ample evidence for this comes from studies using some version of the so-called Visual World Paradigm (VWP), a classic Cognitive-Science paradigm for probing the online integration of linguistic and visual information as indexed by eye movements.\cite{tanenhaus_integration_1995} In a typical VWP study, participants hear an unfolding sentence and see different objects/scenes on a screen,\cite{huettig_using_2011} or even in virtual reality.\cite{eichert_language-driven_2018} Experiments have robustly shown that participants' eye movements index word prediction. For example, just after hearing `the man will eat...', participants look at a cake, rather than non-edible distracters before the continuation of the sentence is uttered.\cite{altmann_incremental_1999} Thus, when visual information is critical for predicting upcoming sentence content, participants tend to direct their eye movements towards the relevant visual information even before hearing the associated word.\cite{altmann_incremental_1999}, making attention to visual cues a cornerstone of predictive language processing in humans.\cite{kamide_integration_2003,kukona_knowing_2014,rommers_verbal_2015}

Recent advances in natural language processing and computer vision have culminated in the transformer architecture, which similarly selects relevant information—linguistic tokens or visual pixels-from the surrounding context, using an `attention' mechanism.\cite{vaswani_attention_2017} This mechanism is not explicitly modeled on human attention, however. For example, transformers have multiple attention heads and can simultaneously attend to many image patches at once, whereas human attention has a more limited capacity,\cite{serrano_is_2019,da_costa_assessing_2020} though it is unclear how many targets human attention can track simultaneously.\cite{busse_spread_2005} Instead, the design considerations for self-attention in transformers are primarily driven by engineering considerations (such as enabling parallel processing) and performance benchmarks against other models.\cite{tay_synthesizer_2021, lindsay_attention_2020} However, there are many emergent capabilities of LLMs that we have not yet understood \cite{wei2022emergentabilitieslargelanguage} and the ability to pay attention to semantically relevant contextual cues in a VWP-type task may be one of them.

Significant overlap between model attention and human eye-tracking would indeed be a surprising finding, as human-model alignment (e.g. between human brain data and vision models) has been said to depend on training diet and less on architectural features \cite{conwell2022billion}. On the other hand, visual transformers exhibit more human-like image classification compared to convolutional neural networks without attention.\cite{dosovitskiy_image_2021} \citet{tuli_are_2021} argue that this fact `could possibly be explained by the nature of attention models that permits focus on the part of the image that is important for the given task and neglect the otherwise noisy background to make predictions'. Interestingly, some models incorporate cross-modal attention mechanisms (e.g. BLIP\cite{li_blip_2022}) which may facilitate semantically guided interactions between visual and textual modalities and therefore present ideal candidates for tracking relevant visual cues during multimodal language processing. 

Bringing these strands together, we assess three related predictions: (1) Adding visual context to the input enables computational models to make predictions that align on average more closely with human predictions. (2) This effect is enhanced for models with attention compared to their counterparts without attention. (3) During the prediction process visual attention patterns correlate with human eye-tracking and multimodal transformer models can attend to semantically salient visual cues.

\section*{Results}

Figure 1 (\textit{Unimodal Paradigm}) illustrates the extraction of predictions from various models and collection of human predictability estimates using audio-based and video-based stimuli. Figure 1 (\textit{Multimodal Paradigm}) presents multimodal model inputs (for models with and without attention) and experiments using audio-visual stimuli. In the first online experiment, 200 participants viewed 100 six-second audio-visual clips from "The Prestige" and "The Usual Suspects". For each stimulus, participants saw a target word and brief instructions (Figure 1a), followed by a six-second clip (Figure 1b and 1e), then rated the clip's relevance for predicting the target word on a 0–100 scale (Figure 1c and 1f). In two follow-up experiments, 200 participants carried out the same task viewing the same clips but without audio, or listening to the audio-only clips respectively. In the two experiments that included visual input, eye gaze was recorded via webcam across all 100 trials.

\newpage
\thispagestyle{empty} 

\begin{figure}[H]
\centering
\includegraphics[width=0.9\linewidth]{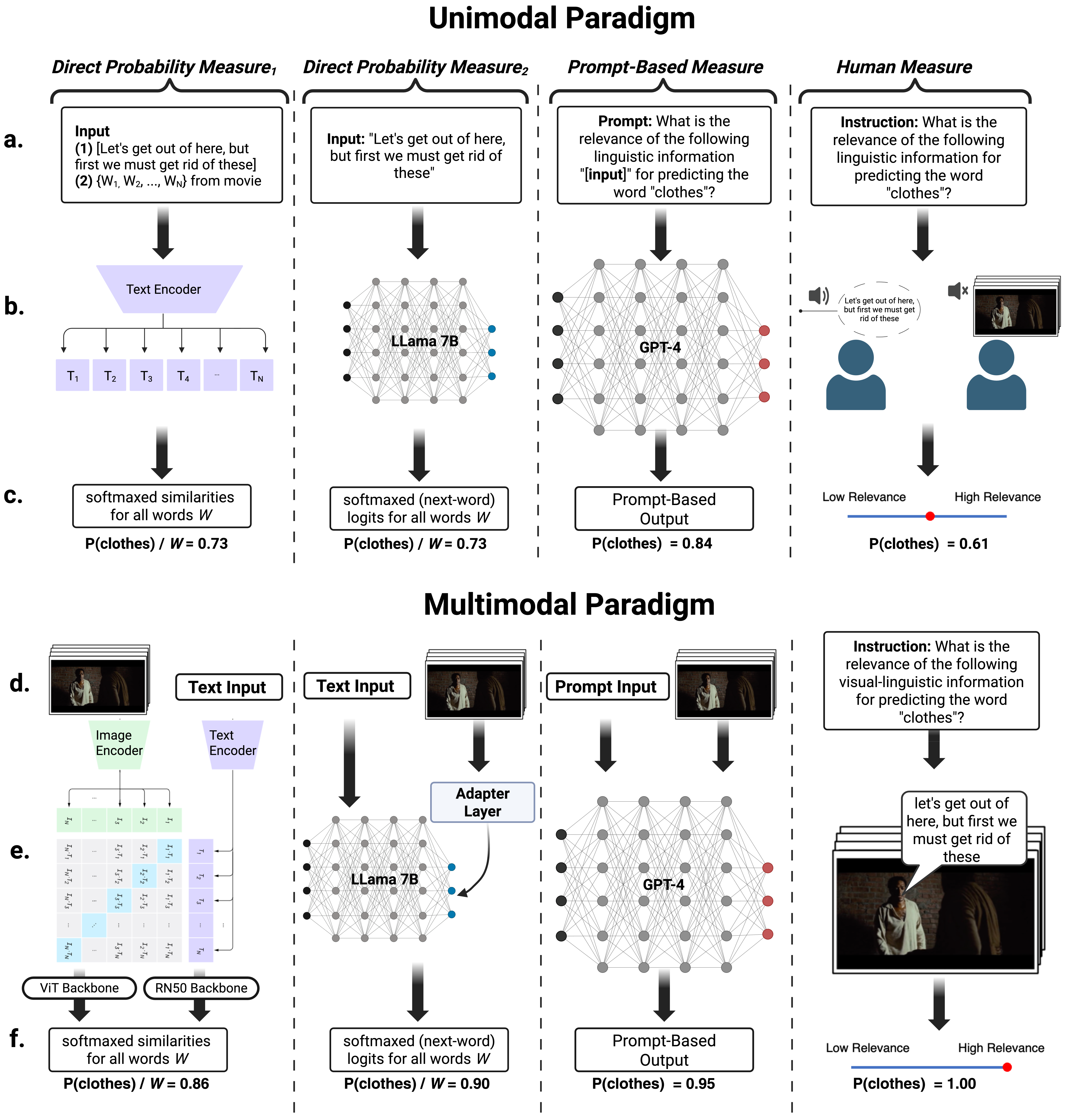}
\caption{\textit{Unimodal Paradigm:} (First column) For the first probability measure, both the incoming dialogue (input) and all labels in the movie (a) are encoded by FLAVA's text encoder in the `text only' input version of the model (b). Predictability is derived as the softmaxed similarity scores (over all labels) between the upcoming label and the resulting encodings (c). (Second column) For the second probability measure, the textual input (a) is fed directly to LLaMA (b). Predictability is derived by pulling out the next-word logits from the model's forward method for all labels in the movie and applying a softmax over them to obtain a probability distribution (c). (Third column) For the prompt-based measure, the textual input is combined with a prompt asking to estimate the predictability of the upcoming word, which is processed by the model (b) and results in a direct prompt-based output measure (c). While GPT-4 and LLaMA are from the same model family, GPT-4 has more parameters than LLaMA. (Fourth Column) For the human measure, instructions are presented to human participants (similar to the prompt used in the prompt-based measure) (a) before they listen to an audio clip (left) or watch a muted video (right) (b) and provide predictability estimates on a Likert scale (c) from 0 (Low Relevance) to 100 (High Relevance). \textit{Multimodal Paraigm:} (First column) For the first direct probability measure, both the incoming dialogue (input) and all labels in the movie (d) are encoded by FLAVA's text encoder. The visual information (frame-by-frame) is encoded by a visual transformer backbone (e). Predictability is derived as the softmaxed similarity scores (over all labels in the movie) between the upcoming label and the resulting multimodal encodings (f). (Second column) For the second direct probability measure, visual input was fed frame-by-frame to the adapter layer. Textual input was fed to the LLaMA model directly (d). Both text and visual information were then processed by the model (e). Predictability scores were derived as the softmaxed next-word logits for all labels in the movie's dialogue (f). (Third column) For the prompt-based measure, the visual input was fed as a GIF to the GPT-4 API, together with a prompt (d). This input was processed by the model with the temperature parameter set to zero (e). Predictability was the direct, deterministic outcome following the prompt (f). (Fourth column) For the human measure, human participants received instructions similar to the prompt fed to GPT-4 in the prompt-based measure (d). Humans then watched the 6 s video clip (e) while their eye movements were tracked through their webcam. Participants indicated relevance on a Likert scale from 0 (Low Relevance) to 100 (High Relevance) (f).}
\label{fig:Figure1}
\end{figure}

\newpage

\subsection*{Prediction 1: Visual Context Aligns Human and Model Predictions}
Our first prediction was that adding visual context to the input enhances behavioural alignment with human language predictions in naturalistic settings. 

We first assessed predictability with two open-source transformer models that share an almost identical inference pipeline. LLaMA-7B \citep{touvron2023llama} is a large-scale generative language model that was trained purely on text; its multimodal variant is obtained by attaching the LLaMA-Adapter visual projection layer \citep{gao2023llamaadapter}. FLAVA-full \citep{singh2022flava} is a 1.3-billion-parameter vision–language encoder that natively accepts both text and images. Despite their architectural differences, both models return token-level logits that can be converted into next-word probabilities.

For unimodal (text-only) evaluation we supplied the subtitle prompt (Figure 1a) and, for every candidate label that occurs in the movie, extracted its probability by applying a softmax over the model’s next-token logits (Figure 1c). For multimodal evaluation we presented the same prompt together with the corresponding video frame: LLaMA receives the frame through the adapter, whereas FLAVA processes it directly (Figures 1d–1e). In both cases we again applied a softmax to the multimodal logits (Figure 1f) to obtain a probability distribution over all candidate labels. 

FLAVA is ideal for comparing unimodal and multimodal prediction, because exactly the same parameters that solve text-only tasks also participate in multimodal reasoning, and vice-versa. Therefore, a single set of decoder weights produces the logits in both settings, so any difference in predictability can be attributed entirely to the additional visual context rather than to architectural or training-objective changes.(FLAVA  performs competitively as a pure language model, attaining a GLUE score of 86.0 \citep{singh2022flava}).  For LLaMA, a potentially confounding factor is that the adapter layer is itself fine-tuned—while the original LLaMA weights remain frozen—on a large instruction-following corpus that pairs CLIP image embeddings with chat-style prompts and textual answers, so part of any multimodal gain may reflect this additional instruction-tuning rather than visual grounding alone.

The extraction of predictability estimates from these models was different to the human experiment: rather than directly probing for next-word prediction, we asked participants to assess the relevance of the visual-linguistic information for predicting an upcoming word. For this reason (and because GPT-4 is a closed model) we tested prediction (1) in a second way with a prompt-based measure that closely resembled the human instructions. At the time the analysis was done, the same GPT-4 model existed in openAI's API in two versions, as text only, and as a text-image model. In the unimodal paradigm, we fed the text model a prompt asking it to estimate the predictability of a given upcoming word, based on provided textual information (Figure 1a). After processing this prompt (Figure 1b) with the temperature set to 0 (to produce deterministic outcomes), GPT-4's output was a predictability estimate (Figure 1c.). In the multimodal paradigm, we fed the same prompt to the text-image model, providing text-based linguistic information in the prompt and the visual content in a Graphics Interchange Format (GIF) (Figure 1d). After processing this information with the temperature set to 0, the output was, again, a deterministic predictability estimate for the upcoming word (Figure 1f). 

We  note that while \citet{hu2023prompting} caution that prompts are not sufficient for showing that a linguistic generalisation is \textit{absent}, temperature-0 prompt-based measures have been shown to provide a reasonably faithful snapshot of the model's next-token probability distribution \cite{mccoy2023embers, renze2024temperature, zhang2024thorough}.

To test whether visual context enhances alignment with human language predictions, we compared the performance of all models in the unimodal and multimodal paradigms. Our findings indicate that alignment with human predictability estimates significantly increased in the multimodal paradigm (Figure 2). The scatter plots in Figure 2 show the predictability estimates of each model against human scores (each data point represents average correlation between all human participants and the model for one of the stimulus clips). Even for a strong model like GPT-4, predictability estimates are biased in the unimodal paradigm (with clusters around 0). Overall, the distributions look more distributed and human-like for all models in the multimodal paradigm. To interpret these model–human correlations, we computed a leave‐one‐participant‐out (L1PO) ceiling: for each of the 200 participants we correlated their clipwise scores with the mean of the remaining 199 participants and averaged the resulting 200 correlations.  The ceiling is $\rho_\text{L1PO}=0.58$.  All ceiling‐corrected values reported below as \(\rho / \rho_\text{ceiling}\) are therefore interpretable as the percentage of explainable variance captured.

Before turning to the behavioural alignment analysis we verified that each architecture’s \emph{text-only} branch is linguistically competent and that any boost observed in the \emph{multimodal} (MM) condition is indeed driven by visual input. Table S5 reports sentence‐level perplexities (\textsc{ppl}) on our stimuli corpus.  For both families the MM variant yields a lower perplexity than the unimodal (U) baseline: FLAVA improves by $4\,\%$ (\textsc{ppl}\,$=205.3\rightarrow196.6$; non-significant, $t=0.80$, $p=.43$) and LLaMA‐Adapter by $5\,\%$ (\textsc{ppl}\,$=510.2\rightarrow484.5$; non-significant, $t=0.93$, $p=0.36$). Crucially, RoBERTa‐base (\textsc{ppl}\,$=142.3$) surpasses \emph{both} multimodal systems, showing that raw language-model fitness is not what drives the visual advantage.  
Together with the “no-collapse’’ result for U branches, this establishes that any gains we later observe in human–model alignment arise from visual grounding rather than from an overall stronger language decoder.

Multimodal FLAVA achieves the strongest raw correlation ($\rho=0.42$), corresponding to $0.42/0.58\approx72\%$ of the explainable variance; GPT-4 multimodal follows at $0.40$ ($68\%$). All unimodal variants explain less than $42\%$ (FLAVA-MM $0.722$, GPT-MM $0.681$, LLaMA-MM $0.273$, FLAVA-U $0.275$, GPT-U $0.413$, LLaMA-U $0.206$). These correlation values are computed after first averaging the responses of all participants for each stimulus clip, so each point in the analysis represents a \emph{single item} (video) rather than an individual trial. In a separate analysis we kept every trial distinct---one row per participant $\times$ video---and assessed prediction with leave-one-video-out cross-validation on the $z$-scored data; under this stricter, single-trial setup the best models yielded $\mathrm{CV}\,R^{2}=0.076$ for FLAVA-MM and $0.061$ for GPT-MM, whereas every unimodal variant produced negative $R^{2}$ (see Supplementary table S).

A linear mixed‐effects model with random intercepts for \emph{Participant} and \emph{Video} confirms this pattern:  
(i)~adding visual context increases the trial-by-trial coupling between model and human scores (\(\beta_{\text{Model}_z\times\text{Modality}} = 0.020,\; z = 4.71,\; p < .001\));  
(ii)~removing visual information decreases the mean prediction down by roughly one standard deviation for both GPT (\(-1.06\) SD) and LLaMA (\(-1.04\) SD) relative to their multimodal counterparts, showing a large calibration penalty;  (iii)~within the multimodal condition FLAVA and GPT exhibit weak but positive slopes (\(\approx +0.006\) SD per SD after interactions are applied), whereas LLaMA retains a small yet significant negative slope (\(-0.029\) SD/SD, \(z=-3.77,\; p<.001\)), indicating residual mis-alignment specific to that architecture. Complete information on all analyses can be found in supplementary Tables S1-S4. 

These results suggest that adding visual context to the input increases behavioral correlation in human and model predictions compared to text-based input. Interestingly, parameter size did not seem to significantly impact alignment with human predictability estimates. The much smaller model FLAVA clearly outperformed the larger LLama and GPT-4 models in the multimodal paradigm. This is surprising given the observed scaling laws in text-only models \cite{kaplan2020scalinglawsneurallanguage} and could suggest a more important role for carefully curated training datasets, specific loss functions, or aspects of model architecture over training diet for multimodal language processing.

\begin{figure}[H]
\centering
\includegraphics[width=1\linewidth]{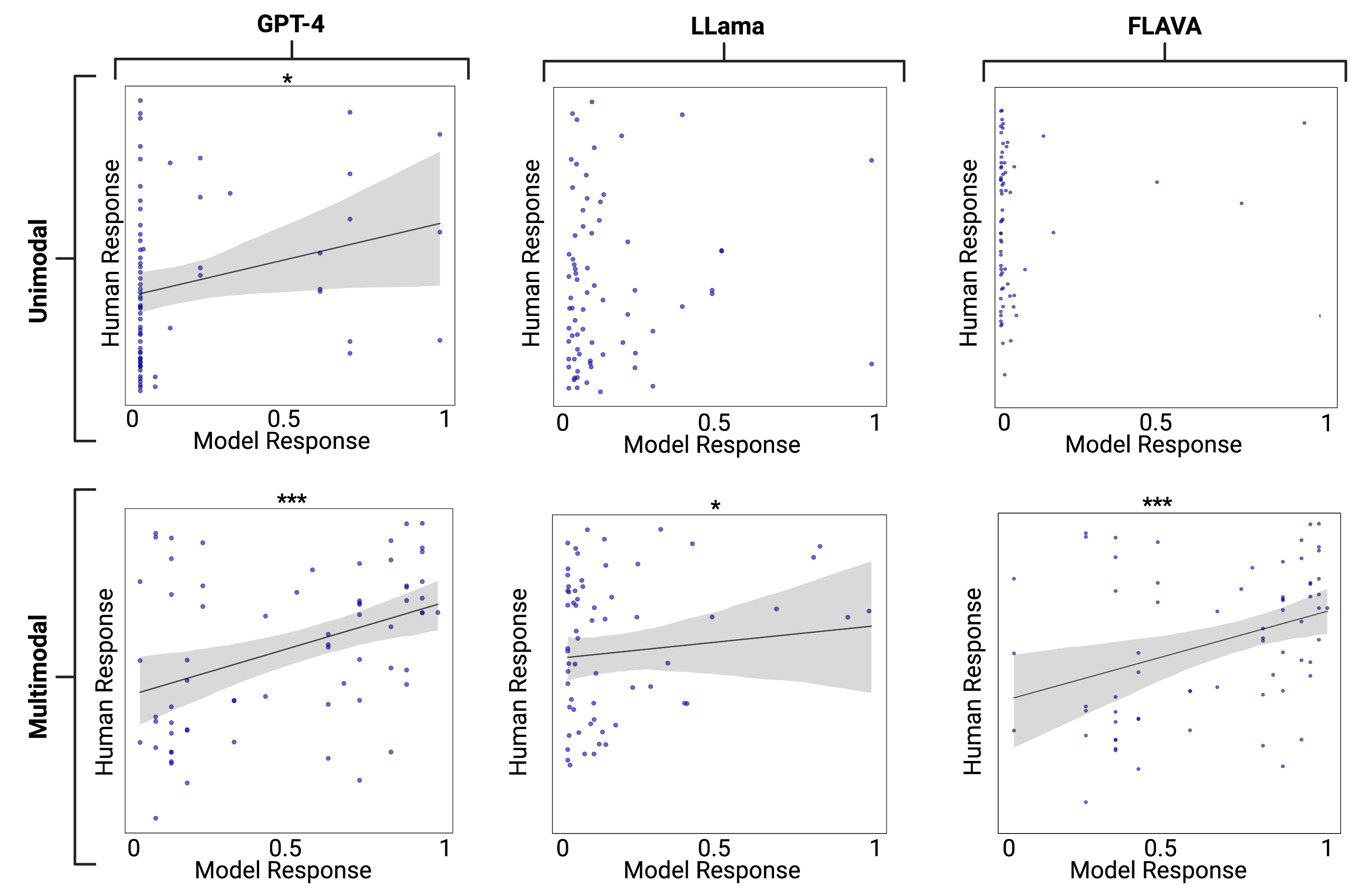}
\caption{Results for comparing model and human predictability scores in unimodal and multimodal paradigms. Average human response per audio-visual (multimodal) and audio-only (unimodal) stimuli (Y-axis) plotted against model response (X-Axis) for unimodal and multimodal GPT-4, LLaMA, and FLAVA. For all multimodal paradigms, the predicted model response (black regression line, only displayed for significant predictions) aligns significantly more with human predictability estimates compared to their unimodal counterpart. All correlations more than 2 SDs from the mean were removed.}
\label{fig:Figure2}
\end{figure}

\subsection*{Prediction 2: Attention matters}
To assess the role of attention mechanisms in aligning model and human predictions in the multimodal paradigm, we used two variants of the CLIP model in the multimodal paradigm. With an estimated 250 million parameters, CLIP \cite{radford_learning_2021} is a relatively small, but highly capable multimodal, contrastive image-language matching model. We compared variants of the CLIP model that incorporate attention mechanisms (those with a Vision Transformer, ViT, backbone) to variants without attention (those with a ResNet, RN, backbone) across different sizes and percentiles of human predictability. 

Higher human predictability is associated with more informative visual cues and therefore constitutes an operationalisation of semantic visual saliency \cite{altmann_incremental_1999}. As CLIP is not designed for next-word prediction, we determined predictability by computing probability scores. We applied the softmax function to the model's similarity assessments between the provided prompts and relevant labels (Figure 1a and Figure 1d) from the movie. This method transforms the model's output into a probability distribution over all possible next words (i.e., next word predictions). 

Permutation tests that contrasted all ViT against all ResNet models revealed a consistent ViT advantage at every percentile of model predictability: at the top 25th percentile (i.e. high human predictability) the difference in correlations was $\Delta r = 0.062$ with $p = 0.007$, at the 50th percentile $\Delta r = 0.089$ with $p < 0.001$, at the 75th percentile $\Delta r = 0.042$ with $p = 0.002$, and at the low 90th percentile $\Delta r = 0.048$ with $p < 0.001$. Bootstrap confidence intervals computed at the individual-observer level showed that ViT-32 remained the strongest single model, achieving $r = 0.396$ (95\% CI [0.358, 0.431]) at the 25th percentile and $r = 0.229$ (95\% CI [0.209, 0.248]) at the 90th percentile, whereas the best-performing ResNet, RN50$\times$64, reached $r = 0.238$ (95\% CI [0.183, 0.289]) and $r = 0.170$ (95\% CI [0.144, 0.196]) at the same percentiles; the weakest ResNet, RN50$\times$4, produced $r = 0.001$ (95\% CI [$-0.057$, 0.063]). These results are summarised in figure \ref{fig:Figure3}. Averaged across percentiles, the three ViT models yielded a mean correlation of $r = 0.217$, whereas the three ResNets averaged $r = 0.150$, giving an overall difference of 0.066 in favour of the attention-based architecture (see figure 3). 

An ordinary-least-squares model fitted to the eight architecture-by-percentile means accounted for 90.6\% of the variance and revealed a strong main effect of architecture ($\beta = 0.216$, $p = 0.005$) together with a negative architecture–percentile interaction ($\beta = -0.0025$, $p = 0.015$), indicating that the ViT advantage is most pronounced when predictability is high (i.e. salient visual information is present). 

Size-matched pairwise regressions confirmed this pattern: for the small comparison (RN50$\times$4 versus ViT-16) the architecture main effect was $\beta = 0.056$ with $p = 0.113$, but the interaction $\beta = -0.099$ with $p = 0.019$ demonstrated a ViT benefit when predictability was high; for the medium model-size comparison (RN50$\times$16 versus ViT-32) the ViT main effect was $\beta = 0.174$ with $p = 0.003$ and the interaction $\beta = -0.102$ with $p = 0.017$; for the large model-size comparison (RN50$\times$64 versus ViT-L) the main effect was marginal ($\beta = 0.045$, $p = 0.055$) and the interaction was non-significant ($p = 0.414$). 

Within-family analyses of parameter count revealed that size improved ResNet performance chiefly at the 25th percentile ($\beta = 0.119$, $p = 0.074$) and became negligible by the 90th percentile ($\beta = 0.017$, $p = 0.621$), whereas size effects among the ViT models were uniformly small and not significant at any percentile (largest $\beta = 0.103$, all $p > 0.37$). 

A comprehensive regression that included architecture, size rank, percentile and all interactions explained 62.8\% of the variance (overall $F(7, 16) = 3.85$, $p = 0.012$) and confirmed that the principal predictors of human–model agreement are percentile, model size and the architecture–percentile interaction, with no evidence that larger ViT models derive disproportionately more benefit from additional parameters than similarly sized ResNets. 

In order to corroborate these results, we conducted attention ablation experiments using a single transformer backbone (CLIP ViT-B/32). We compared the standard learned attention patterns against two ablated variants: uniform attention (where all spatial locations receive equal attention weights) and patch-shuffled attention (where learned attention weights are randomly redistributed across spatial locations). Mixed-effects regression analysis showed that both ablation conditions produced identical disruptions to human-model alignment. While learned attention showed a positive relationship between model confidence and human ratings ($\beta = 0.063$, $p < 0.001$), both uniform and patch-shuffled attention exhibited significant negative interactions ($\beta = -0.196$, $p < 0.001$ for both conditions), effectively reversing the model-human correlation. Specifically, under learned attention, higher model confidence predicted higher human ratings, whereas under both ablated conditions, higher model confidence predicted lower human ratings (for full model output see supplementary table S6). The identical effects of uniform and shuffled attention suggest that successful human-model alignment depends not merely on attention variability or spatial organization, but on the specific learned content of attention patterns.

Taken together, these analyses indicate that attention mechanisms enable CLIP ViT models to track inter-individual variability in human judgments more effectively than ResNet backbones, particularly when relevant visual information is present; detailed permutation, bootstrap and regression outputs for every model, percentile and comparison are provided in Supplementary Tables S6-S13.

While higher predictability percentiles contain more informative visual information for human participants, it is not clear to what extent the model is really attending to them. Therefore, we next sought to understand if, when, and why human visual attention, as indexed by eye gaze, overlaps with the model's visual attention in our multimodal paradigm.

\begin{figure}[H]
\centering
\includegraphics[width=0.8\linewidth]{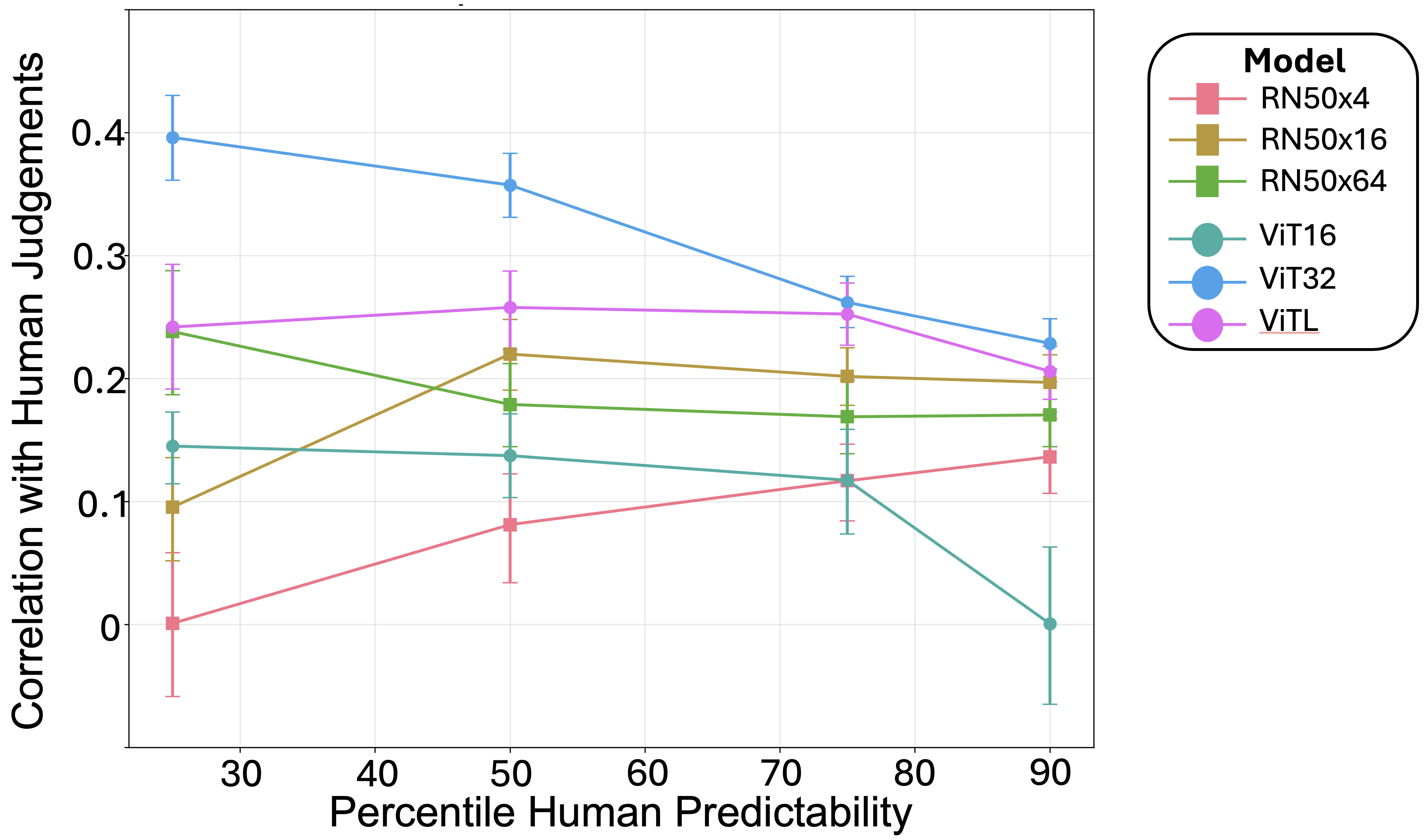}
\caption{Individual‐level comparison of Pearson correlations between human relevance judgments and model responses across percentile thresholds of model predictability for six CLIP architectures (RN50x4, RN50x16, RN50x64, ViT16, ViT32, ViTL; lower percentile means higher predictability). Each marker shows the mean correlation at the 25\textsuperscript{th}, 50\textsuperscript{th}, 75\textsuperscript{th}, and 90\textsuperscript{th} percentiles of the model’s output distribution, with vertical bars indicating 95\,\% bootstrap confidence intervals (1\,000 resamples). Square markers denote ResNet variants; circles denote Vision Transformer (ViT) variants.}
\end{figure}

To test whether visual input suppresses predictive uncertainty, we compute a per-sentence entropy drop for each open architecture. For every sentence \(s\), we estimate a text-only entropy \(H_{\text{text}}(s)\) and a multimodal entropy \(H_{\text{mm}}(s,i)\) for each of the 15 frames, then take the best image \(i^\star\) with the lowest entropy and report
\begin{equation}
\Delta H(s)
=
\max\!\bigl(0,\, H_{\text{text}}(s) - H_{\text{mm}}(s,i^\star)\bigr).
\end{equation}
Raw, unclamped values are also logged.

Using this procedure, we observe reliable positive reductions relative to text-only: FLAVA \(\overline{\Delta H}_{\text{FLAVA}} = 0.26\) bits; LLaMA-adapter \(\overline{\Delta H}_{\text{LLaMA}} \approx 0.17\) bits; and BLIP \(\overline{\Delta H}_{\text{BLIP}} = 0.24\) bits.

As an auxiliary check on visual informativeness, we compute for CLIP the entropy drop of the image-set distribution \(p(\text{image}\mid\text{text})\) relative to a uniform baseline,
\begin{equation}
\Delta H_{\text{CLIP}}
=
\log_2(15) - H\bigl(p(\text{image}\mid\text{text})\bigr).
\end{equation}
This quantity is non-zero on average in this run, indicating markedly non-uniform (“peaky”) evidence across frames.

Together with the gaze–attention findings, these results indicate that visual cues can constrain a model’s predictive distribution.

\subsection*{Prediction 3: Overlap Between Model Attention and Human Eye Tracking}
To this end, we tested our third prediction by directly comparing attention weights from two models'  attention layers to human eye tracking data collected during the online task. Although we collected output from all layers, the late-intermediate layers of transformer models are thought to synthesise the earlier layers' processing into a comprehensive output.\cite{tenney_bert_2019, rogers_primer_2021} For example, previous studies suggest that the best semantic features for predicting brain responses to natural language can be extracted from late-intermediate layers (especially layers 9 and 10 in models like GPT-2, which usually have around 12 layers).\cite{liu_linguistic_2019, oota_joint_2022, caucheteux_brains_2022, jain_computational_2023, tang_semantic_2022} If models attend to semantically salient visual features, then we should expect attention weights from these layers to correlate most significantly with human eye tracking. 

We compared three distinct types of attention mechanisms: CLIP's visual attention, which is shaped by contrastive image-text pretraining, BLIP's pure visual attention, which operates independently of language input, and BLIP's cross-attention, which explicitly models interactions between visual and linguistic features.  

To compare each model's attention matrices to human attention data from eye tracking, we analysed the attention matrices for each of the frames from the video clips used in our experiment. In order to isolate purely visual attention, we also recorded eye-tracking of 200 participants who had to estimate predictability from the visual input alone (rather than the video-speech input). In both cases, matrices represent pixel importance in predictability scores for the model and the fixation duration for humans. We averaged these matrices over 400ms segments, considering the natural latency in saccades (250–400ms),\cite{bergen_experimental_2007, fischer_express_1993} the relevance of this time frame for predictive language processing studies using EEG,\cite{dimigen_auditory_2022, kutas_thirty_2011} and the average word length in English being around 400ms.\cite{marslen-wilson_linguistic_1973} Unlike the widespread distribution of visual transformer attention, human attention tends to be more focused. We therefore adapted our analysis by thresholding the model's attention and applying Gaussian smoothing to both sets of data, choosing parameters to highlight differences between actual human attention and a random distribution. We then created probability distributions from these heatmaps and quantified alignment using Spearman correlation. A human ceiling correlation value for each of the 15 heatmaps per video clip was determined by correlating each participant's probability distribution with the $N-1$ (199) other probability distributions for this heatmap and taking the average of these correlations for each of the 15 segments per video clip. 

We found that attention patterns from BLIP's cross-attention layers showed the highest alignment with human eye-tracking data from our main task, explaining 70\% of the variance in the human ceiling in layer 9 (see Figure 4(b)). In contrast, CLIP's attention patterns showed lower alignment, explaining 52\% of the variance in the human ceiling. In the purely visual task, the human ceiling was much lower, but so was the model-human alignment, explaining 45.4\% of human variance. 

There are several potential reasons for the higher correlation observed in BLIP's cross-attention layers. Cross-attention mechanisms allow the model to directly integrate linguistic context into its visual attention\cite{li_blip_2022} Additionally, the late-intermediate layers (layers 9 and 10) may capture higher-level semantic representations that are more aligned with human attention patterns. These layers likely integrate complex features such as objects, actions, and settings, which are crucial for understanding and predicting language in dynamic visual contexts. CLIP on the other hand encodes both modalities (image and text) separately, so visual attention is computed taking language into account only to the extent that the visual encoder is biased towards language by the contrastive training task of matching images and captions. Therefore, the fact that we \textit{do} find overlap is already surprising and we do not expect a close to perfect overlap. These findings therefore suggest that explicitly modeling the interaction between visual and linguistic features, rather than relying on joint encoding or purely visual attention, may be crucial for better performance in multimodal language prediction tasks.

\begin{figure}[H]
\centering
\includegraphics[width=1\linewidth]{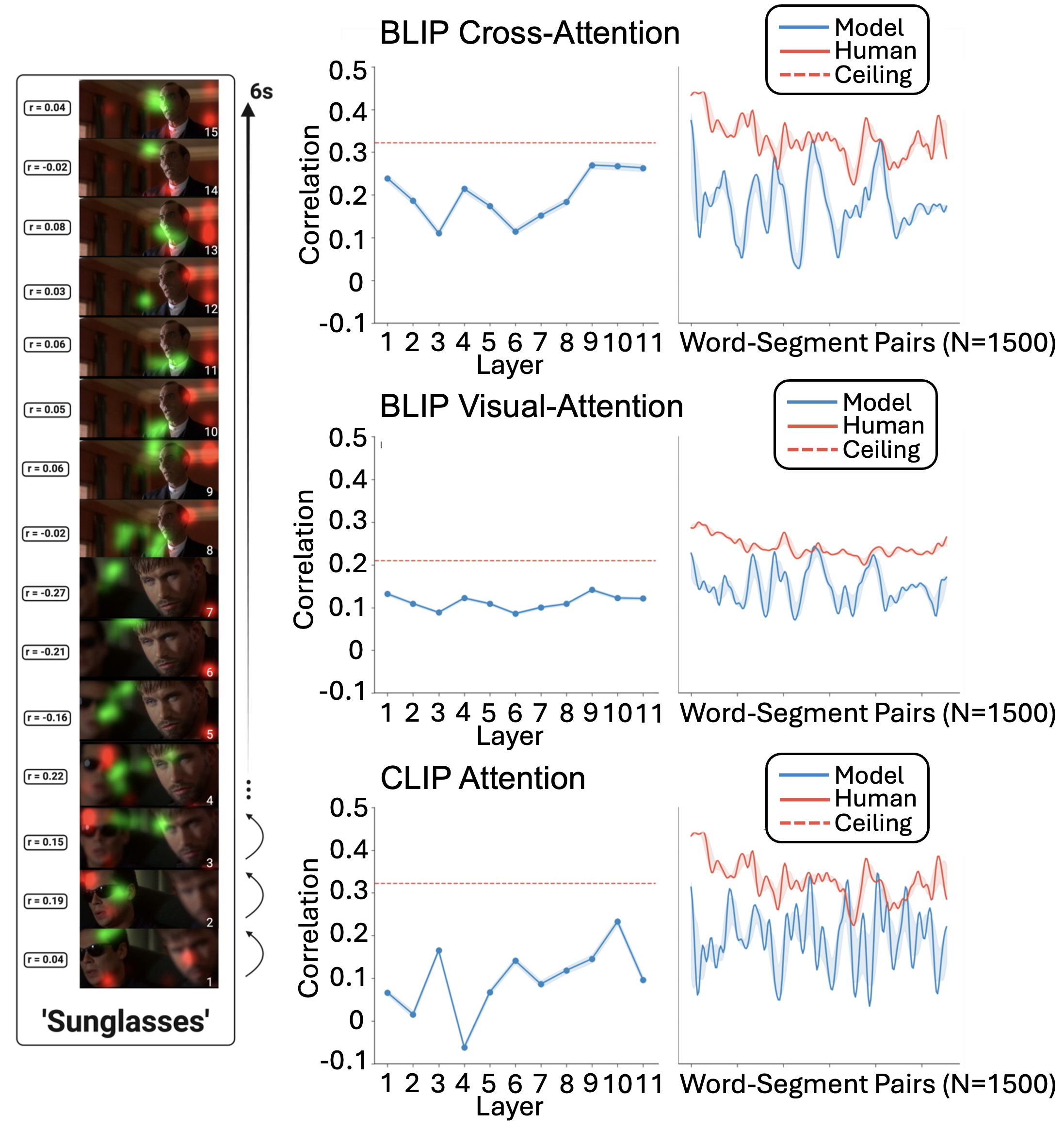}
\caption{Alignment between human gaze (green) and model attention (red). (left) One example of evolving alignment in layer 9 between model attention patterns (red) and human eye tracking (green) over one video clip, separated into 15 segments. When salient visual information was present (segments 2, 3, and 4), correlation was positive. However, when the referent was not present, correlation dipped into negative values (e.g. segment 5, $\rho$ = -0.16). (right) Layerwise  alignment between model attention and human eyetracking. Average alignment for all segments per layer is displayed as a blue line (with blue dots) on the left. Alignment for each segment for the layer with best fit is displayed as the blue line on the right. The human ceiling is displayed as red line, with a mean inter-human correlation of 0.32. For BLIP cross-attention, the layer with the best fit was number 9, with a mean r of 0.27, which explained 70\% of the variance of the human ceiling. For BLIP with visual attention only, the mean r was 0.142, which explained 45\% of the variance of the human ceiling. For CLIP, the layer with the best fit was number 10, with a mean r of 0.233, which explained 52\% of the variance of the human ceiling.}
\label{fig:Figure5}
\end{figure}

\subsection*{Semantic Characteristics of Attention Alignment}
To better understand when and why model attention and human eye-tracking aligned, we conducted several follow-up analyses. First, we evaluated our findings using multiple established saliency metrics. We used similarity score (SS) to assess the overall spatial distribution match between model and human attention maps, normalized scanpath saliency (NSS) to evaluate how well the model predicts human fixation locations while controlling for center bias, and information gain (IG) to measure the added value of model predictions over a center bias baseline \cite{bylinskii_what_2019, kuemmerer_saliency_2018}. BLIP's cross-attention consistently outperformed other attention mechanisms across all metrics (SS: 0.71 ± 0.04; NSS: 2.14 ± 0.12; IG: 0.89 ± 0.06), followed by CLIP's attention (SS: 0.54 ± 0.05; NSS: 1.82 ± 0.14; IG: 0.67 ± 0.07), while BLIP's visual attention showed the weakest alignment (SS: 0.42 ± 0.06; NSS: 1.45 ± 0.15; IG: 0.51 ± 0.08). These results indicate that the superior performance of BLIP's cross-attention is robust across different evaluation criteria and cannot be explained by measurement artifacts or center bias.

To investigate the semantic basis of these alignments, we conducted a detailed annotation study using representational similarity analysis (RSA). Each segment (N=1500) was categorised by twenty-five participants based on the presence of predictive visual information: direct referents, associated objects, or broader contextual cues (e.g., settings, actions, people, or emotions) that could facilitate word prediction. This approach builds on established frameworks for analyzing predictive processing in naturalistic contexts \cite{bar2021objects, henderson_meaning_2017}. We constructed representational dissimilarity matrices (RDMs) from participants' rating patterns and compared these with model RDMs derived from layer-wise feature representations using Spearman rank correlations. Representational similarity analysis revealed systematic patterns of semantic emergence across transformer layers, with BLIP demonstrating consistently superior representational alignment with human semantic judgments compared to CLIP (see figure 5). 

\begin{figure}[H]
\centering
\includegraphics[width=0.7\linewidth]{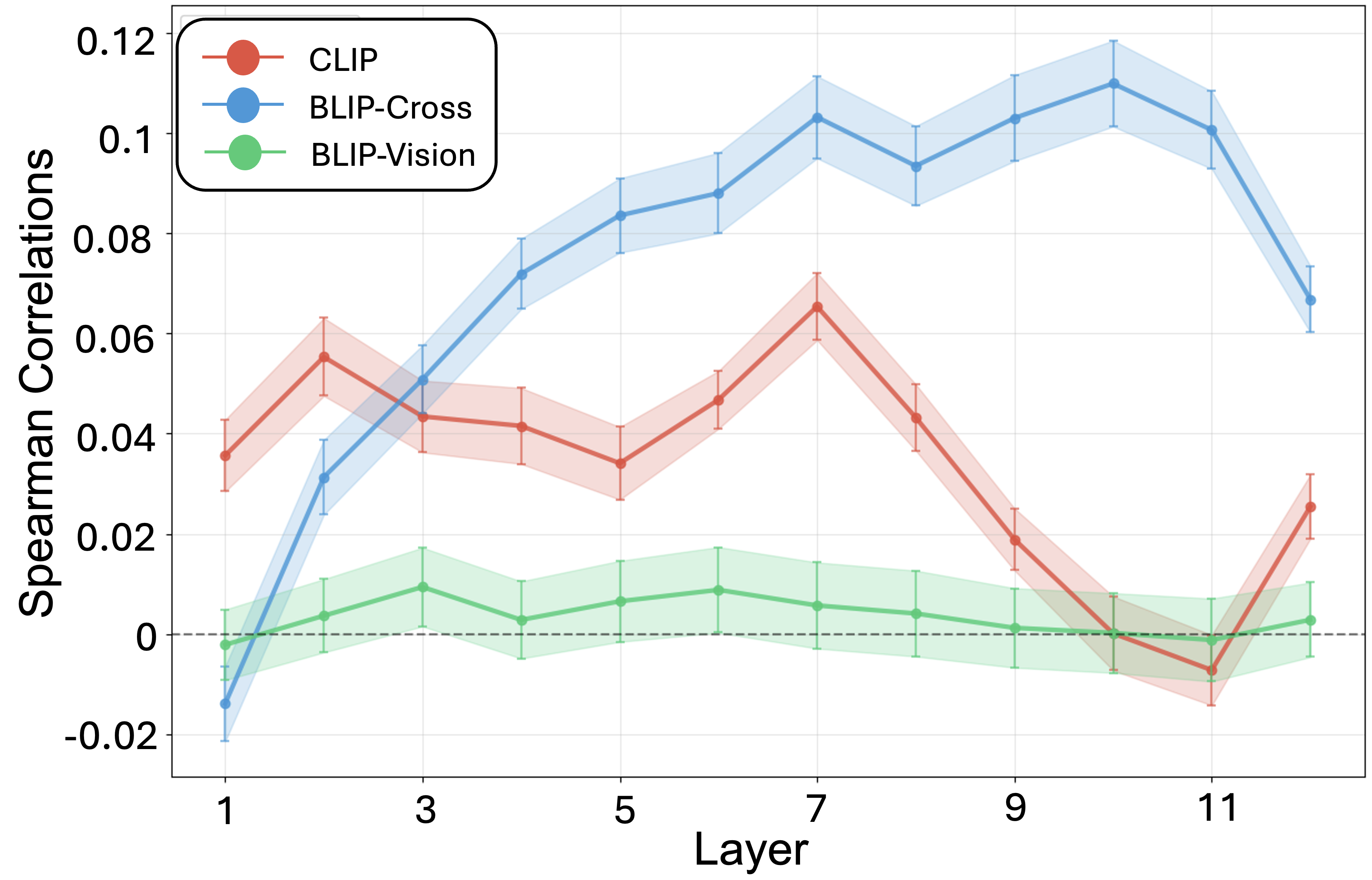}
\caption{Layer‐wise representational alignment between model embeddings and human semantic judgments.
Mean Spearman correlations ($\pm$\,95\% bootstrap CI across 1{,}500 video segments) between human representational dissimilarity matrices (RDMs) and RDMs derived from successive transformer blocks of each model.  BLIP’s cross‑modal fusion stream (blue) exhibits a monotonic increase that peaks around layer~10 ($r\approx0.12$), significantly outperforming both CLIP’s visual encoder (red; peak $r\approx0.07$) and BLIP’s unimodal vision stream (green; $r<0.03$ at all layers; all $p<.001$, FDR‑corrected).  Shaded regions denote $\pm$\,1\,SEM, and the grey dashed line marks the chance level obtained from 10{,}000 layer‑wise label shuffles.  These results indicate an emergent semantic representational structure that is strongest in BLIP’s cross‑attention pathway and only weakly present in purely visual streams.}
\end{figure}

Interestingly, the degree of overlap between the model's spatial attention and human gaze within a segment was modestly correlated with the contribution of that segment to the model-human semantic alignment, but this dependency was modulated by both architecture and the temporal direction of the attention trajectory (Figure 6).  
We first classified every word–segment pair according to whether the frame‑wise attention–gaze correlation increased (rising) or decreased (falling) over the 6s clip.  We then computed a simple, segment‑level Spearman correlation between attention–gaze overlap and the segment’s contribution to the global RSA score.  
For rising segments (green), BLIP’s cross‑modal stream showed a modest yet significant positive association (ρ=0.164, p<.01), whereas the relationships for CLIP (ρ=–0.018, n.s.) and BLIP’s unimodal vision stream (ρ=–0.029, n.s.) were negligible. In contrast, for falling segments (red) the pattern flipped: both CLIP (ρ=–0.101, p<01) and BLIP‑Vision (ρ=–0.136, p<.001) exhibited reliable negative relationships, while BLIP‑Cross  showed no significant effect (ρ=–0.041, n.s.).  

Taken together, these results indicate that BLIP’s cross‑modal mechanism benefits when its attention increasingly converges on the locations humans will ultimately fixate, boosting semantic alignment.  By comparison, in purely visual streams a loss of alignment over time is associated with poorer semantic representations. This emphasises the importance of sustained, semantically driven attention in multimodal processing.

\begin{figure}[H]
\centering
\includegraphics[width=1\linewidth]{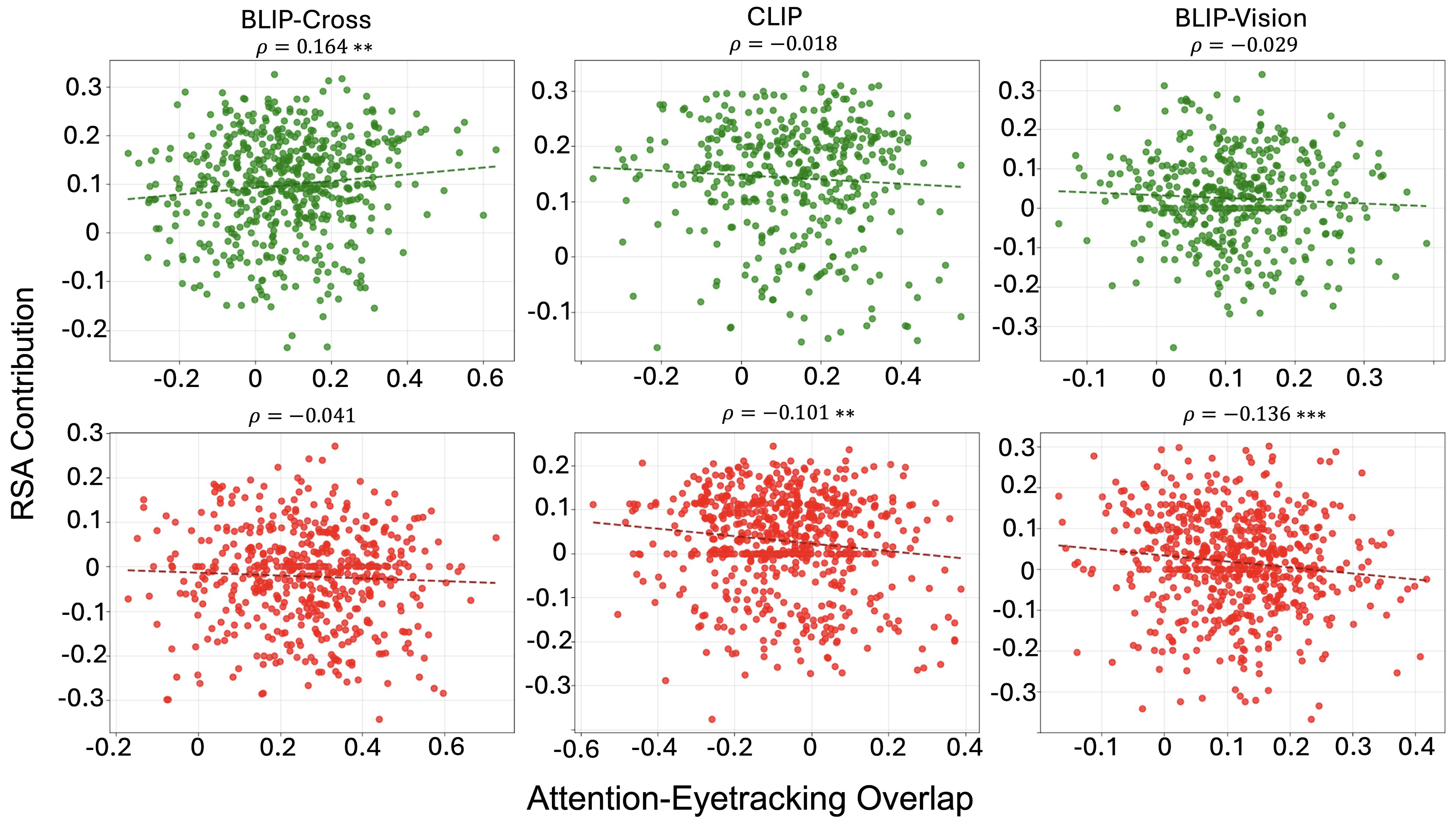}
\caption{Segment‑level link between attention–gaze overlap and semantic contribution.
Scatterplots show, for each model, the relationship between a segment’s attention–eyetracking correlation (horizontal axis) and its contribution to the overall RSA score (vertical axis).  
Top row (green) depicts segments in which the attention–gaze correlation increased over time (rising trajectory); bottom row (red) shows segments in which it decreased (falling trajectory).  
Dashed lines are robust linear fits; numbers above each panel report the Spearman correlation ($\rho$). Only BLIP‑Cross exhibits a modest positive relationship under rising trajectories ($\rho$ = 0.164**), whereas CLIP and BLIP‑Vision display significant negative relationships under falling trajectories ($\rho$ = –0.101**, $\rho$ = –0.136***, respectively).  Axes labels are shared across panels for comparability.} 
\end{figure}

\section*{Discussion}
We tested the same pretrained neural networks in two input paradigms—unimodal (text only) versus multimodal (text\,+\,image)—within a large-scale, web-based Visual World Paradigm. Our analyses reveal several key results. First, adding visual context to the input increases behavioural correlation between model and human language predictions. Second, model size is not everything: depending on their architecture, low-parameter models align with human predictability estimates better than models with orders of magnitude more parameters. Third, attention matters: CLIP performed significantly better with a transformer compared to a resnet visual backbone. Fourth, transformer attention patterns show substantial overlap with human eye movements when semantically relevant visual cues are present. The remainder of the discussion unpacks each of these findings in turn.

The correlation and R² values we observe between human and model predictions deserve careful interpretation within the broader context of psycholinguistic research. While the absolute magnitude of our best cross-validation R² values (0.076 for FLAVA-MM and 0.061 for GPT-MM) might appear modest, they are comparable to values typically reported in state-of-the-art psycholinguistic modeling of human language processing. For instance, studies predicting reading times based on surprisal metrics often report R² values between 0.05-0.10 \citep{smith2013effect}, and those modeling neural responses to speech from predictability measures typically achieve R² values of 0.06-0.10 \citep{broderick2018electrophysiological}. The mixed-effects model results further support the importance of multimodal context, with significant interactions between modality and model predictors. Notably, the robust negative correlation between the unimodal variants is consistent with findings from \citet{frank2015erp}, where higher surprisal in unimodal settings often corresponds to increased processing difficulty. Our video-level correlations, which average across individual trials, show stronger alignment (ρ=0.42 for FLAVA-MM) and highlight how aggregation can reveal clearer patterns by reducing trial-level noise—a methodological consideration echoed in related work on multimodal comprehension \citep{kuperberg2020neural}. These values become even more meaningful when considered as a proportion of explainable variance (72\% for FLAVA-MM), suggesting that our best multimodal models capture a substantial portion of the systematic variance in human predictions that can theoretically be explained after accounting for participant and stimulus noise.

From the standpoint of cognitive science, it may come as no surprise that adding visual context aids prediction. The influential framework of active inference, for example, views the brain as constantly trying to resolve uncertainty \cite{Friston2017Active}, especially when processing language in naturalistic settings with multimodal context.\cite{Friston2015ActiveCommunication} The more information, the better. 

However, brains have evolved over millions of years within a noisy, multimodal environment, learning to map different conceptual modalities onto each other.\cite{Gentner1982StructureMapping} While it may be true that deep neural nets are the right kind of machinery to achieve the same fit to nature,\cite{hasson_direct_2020} our results suggest (contrary to a long strand of literature in cognitive science \cite{Fodor1975LanguageOfThought, Landauer1997LSA, Mikolov2013WordEmbeddings, MahonCaramazza2008EmbodiedCritique, Mahon2015WhatIsEmbodied, FodorPylyshyn1988ConnectionismCritique, Jackendoff2002Foundations}) that language alone may not be an effective `window' to the kind of grounding neural networks would receive from embodied `experience'. 

This is evidenced by the fact that learned integration of modalities, rather than parameter count, was critical for behavioural correlations with human predictive processing. Even though GPT-4 has learnt orders of magnitude more parameters than FLAVA, there is no significant difference in alignment with human scores during our VWP task (in fact FLAVA performed better). We suggest that this is because the GPT family of models was originally trained on language only, while FLAVA has learned to associate language with imagery from scratch. This fits with emerging evidence that effective multimodal processing depends more on architectural constraints than model scale \cite{raffel2020exploring, jaegle2021perceiver} and, as an interpretation of our results, aligns with recent work on the importance of structural alignment in multimodal learning \cite{tsimpoukelli2021multimodal, alayrac2022flava, jaegle2021perceiver}. Instilling multimodal knowledge post-hoc in a language model may not be as effective as jointly learning modalities from the scratch -an idea that is made explicit in the proposal of an embodied' Turing test \cite{zador_catalyzing_2023}.

However, it is not clear whether structural alignment could \textit{never} happen post-hoc. Linguistic data may encode rich multimodal relationships at an abstract level \cite{andreas2022language}, yet alignment could ultimately depend on the availability and quality of large-scale multimodal data that match or exceed the abundance of linguistic training data.\cite{bommasani2021opportunities} Such multimodal datasets are currently not available. However, incorporating sensory data from robotic agents into an existing LLM could constitute a promising way forward. For example, the PaLM-E model \cite{noauthor_palm-e_nodate} extends beyond traditional text and image processing, engaging with real-world robot states and environmental interactions. This approach offers a glimpse into future AI capabilities where computational models may process a richer array of sensory inputs, akin to human perception. PaLM-E's state-of-the-art ability to adapt across various robotic platforms and handle different modalities like images and robot states exemplifies the potential of AI systems to structurally align different modalities and, as a result, operate in more complex, real-world scenarios. Future research continuing on this path may offer an alternative route to bottom-up engineering of human-like AI. 

Whether incorporating different modalities post-hoc or training multimodal models from scratch, it is critical to understand what kind of architectures support effective multimodal integration (for a review on this topic see \citet{baltrusaitis2019multimodal}). Transformer attention is one of the most impactful architectural innovations of the last decades. Our results suggest that it played an important role in model-human behavioural correlation, especially in the more predictable percentiles of the multimodal paradigm. The superior performance of attention-based architectures can likely be attributed to the following factors. 

Compared to CNNs, which are constrained by local processing and fixed structural biases \cite{battaglia2018relational}, transformer-based models with attention can dynamically adapt their focus based on task demands \cite{dosovitskiy_image_2021}. This flexibility appears particularly important for processing naturalistic stimuli, where relevant information may be distributed across the input \cite{wang2023attention}. Furthermore, attention mechanisms, especially cross-modal attention as implemented in BLIP, allow for more sophisticated integration of visual and linguistic information \cite{tan2023survey, xu2023understanding}.

Importantly, the strength of attention-based behavioural correlation varies systematically with the presence of predictive cues in the input. Research on eye tracking during naturalistic language processing suggests that visual attention most reliably indexes predictive processing when relevant cues are situated in a joint attention space \cite{heyselaar_we_2020}. When predictive visual cues are absent, humans show greater variability in what they consider relevant for visual-linguistic integration and prediction \cite{ryskin_prediction_2023}, leading to more diverse judgments and weaker within-human correlations. We observe a parallel pattern in model behavior: CLIP's attention patterns become more diffuse when salient visual cues are absent, resulting in increased variance in predictability estimates and reduced human-model alignment. 

Trying to understand the systematic variance of attention and human eye-tracking overlap, our results showed that BLIP with only visual input did not exhibit strong overlap patterns for any predictive visual cues, which highlights the importance of cross-modal integration in matching human attention patterns \cite{wang_aligning_2023}. In contrast, BLIP's cross-attention showed significantly stronger correlation with human eye-tracking when salient visual cues were present, particularly for high-level semantic features such as setting, actions, people, or emotional content \cite{li2022blip, dou_coarse--fine_2022}. This superior alignment likely stems from BLIP's direct integration of linguistic information into visual attention, enabling more effective exploitation of predictive associations between modalities \cite{custers_learning_2011, 10.1145/3503161.3548283, conwell2024usingmultimodaldeepneural}. CLIP's attention patterns showed strongest overlap specifically for concrete objects or referents \cite{singh_know_2022}, potentially reflecting its training objective of matching specific image regions to textual descriptions \cite{radford2021learning}. While these claims remain speculative, the distinct patterns overall suggest that different attention mechanisms correlate with different and potentially complementary aspects of human visual processing \cite{zhang2023emergent}.

To better understand these effects, we conducted a layer-wise analysis of attention patterns throughout the models. This analysis revealed that the strongest correlations with human eye-tracking emerged in intermediate to late layers (particularly layers 9 and 10). We speculate that the prominence of layers 9 and 10 may reflect their position in a processing hierarchy that has been described mechanistically before \cite{abdullah2023understanding, xu2023understanding, yang2020semantichierarchyemergesdeep}. On this understanding, layers 9 and 10 are early enough to maintain spatial information necessary for attention alignment, yet deep enough to have developed sophisticated semantic representations through successive transformations of the input. This interpretation aligns with previous findings that late-intermediate transformer layers are optimal for extracting semantic features that predict human brain responses during language processing.\cite{caucheteux_brains_2022}.

A representational similarity analysis (RSA) further corroborates these findings by linking the quality of a model’s internal representation to the spatial deployment of its attention.  Specifically, we looked at segment‑wise RSA scores and asked whether segments that sharpen the model–human representational match are the same segments in which the model “looks where humans look”.  As shown in Figure 6, this is indeed the case for BLIP’s cross‑modal version: segments that boost RSA also display stronger, rising attention–gaze overlap.  Conversely, in purely visual streams (CLIP and BLIP‑Vision) the relationship either disappears or inverts, with segments that lower the representational match tending to exhibit declining overlap.  These converging results suggest that cross‑modal attention not only aligns with where humans look, but does so because it taps into the same high‑level semantic regularities that guide human prediction in naturalistic scenes.

Our findings are a first step in understanding the role of transformer attention in multimodal integration and raise important questions about the architectural requirements for human-like attention in multimodal models. While our results demonstrate that cross-modal attention mechanisms can achieve significant behavioural alignment with human attention patterns, particularly in the presence of predictive visual cues (Figure 4(c)), they also reveal important differences in how models handle uncertainty. BLIP's cross-attention shows human-like centralising behavior when obvious visual cues are absent  - a documented effect in the visual world paradigm related to efficient allocation of attentional resources \cite{magnuson_fixations_2019, foulsham_where_2011}. CLIP's tendency to focus on peripheral areas in these situations suggests that achieving human-like uncertainty handling may require architectural innovations beyond current attention mechanisms. Future work might explore whether alternative architectures, such as hierarchical attention networks or dynamic routing mechanisms, could better capture the flexibility and efficiency of human attention allocation across different predictive contexts.

Language based models have celebrated great successes over the past three years, but recently seem to have hit a point of diminishing returns. Our study suggests that adding modalities to the embedding space may be an important next step. Here, it is key to understand how models process and integrate these modalities with language. We suggest that attention (and in particular cross-attention) may be an important factor for driving more human-like model predictions.  

We speculate that the superior performance in the multimodal paradigm of models with (cross-modal) attention could suggest that such architectural features may be key to the neural mechanisms underlying the behavioural regularities we report here \cite{goh2021multimodal}. Recent studies have already linked transformer embeddings to brain responses during unimodal language processing \cite{schrimpf_neural_2021,caucheteux_brains_2022}; our results imply that this correspondence could become even stronger when the embeddings are derived from multimodal models.

\section*{Methods}
\subsection*{Online behavioural studies}

We conducted a series of behavioral and computational experiments to investigate how humans and AI models process multimodal information during naturalistic language prediction. Our approach combined three key components: (1) online behavioral studies where participants viewed movie clips and predicted upcoming words, (2) computational modeling using state-of-the-art multimodal AI systems, and (3) detailed comparison of human and machine attention patterns through eye-tracking data. To isolate the contribution of different modalities, we collected data in three conditions: audiovisual (N=200), audio-only (N=200), and visual-only (N=200). Stimuli consisted of 100 six-second clips from two movies ("The Prestige" and "The Usual Suspects"), carefully selected to represent varying degrees of visual relevance for word prediction. On the computational side, we analysed several models with different architectural features: CLIP (both ViT and ResNet variants), BLIP, LLaMA with visual adapter, GPT-4, and FLAVA. A follow-up semantic annotation study (N=250) provided detailed categorization of the predictive visual features present in the stimuli. This comprehensive approach allowed us to assess not only whether models make similar predictions to humans, but also whether they achieve these predictions through similar attentional mechanisms.

Overall we recruited participants for four studies: first our audio-visual study (n = 200), second our study with audio-only stimuli (n = 200), third our follow-up study with vision-only stimuli, (n = 200), and finally our visual annotation study, (n = 250). Recruitment for all studies worked in the same way via Prolific \cite{palan_prolificacsubject_2018}(https://www.prolific.co). To be included in the study, participants had to be aged 18-50 (inclusive) and fluent in English (having spoken English regularly for at least the past 5-10 years). Inclusion criteria included giving consent to have their camera on as well as having their eyes tracked for the duration of the experiment. We also excluded any participants from participating who had previously seen either of our two movies. All participants finished the study and produced usable data, none had to be excluded post data collection. The final participants for the main study (n=200) were 43\% females (n = 86) with mean age of 33.88 years (range = 23-48 yrs, SD = 13.74 yrs). Countries of origin were the UK (n = 123) and the US (n = 77). The final participants for the visual follow up study (n=200) were 46\% female (n = 92) with mean age of 29.8 (range = 21 - 42 yrs, SD = 11.37). Countries of origin were the UK (n = 134) and the US (n = 66). The final participants for the audio follow up study (n=200) were 42\% female (n = 84) with mean age of 30.5 (range = 21 - 47 yrs, SD = 12.19). Countries of origin were the UK (n = 105) and the US (n = 95). Participants in all studies were paid £9.50/hour. All studies were approved by the UCL Ethics Committee. Additionally, all participants provided written informed consent before participating in the experiment. 

100, 6s movie clips from the films ‘The Usual Suspects’ and ‘The Prestige’ were chosen respectively. All words in these movies were previously annotated using automated approaches with a machine learning based speech-to-text transcription tool from Amazon Web Services (AWS; https://aws.amazon.com/transcribe/) and later corrected by human annotators. This alignment between text and speech meant that the model and humans really received the same dialogue-based information.

To choose the words to be predicted by human participants in our study and later by the model,  all function words (as defined in this list: https://semanticsimilarity.wordpress.com/function-word-lists/) from the movies were excluded and the 6s scenes leading up to but excluding each word were extracted – this amounted to a few thousand words per movie on average. A length of 6s was chosen, as this constituted a good balance between containing enough information while not being too confusing. Furthermore, this meant that it was possible to conduct a large number of trials without causing too much fatigue in participants. For our final study, only  50 scenes were included from each movie, leading to 100 clips, totalling 600s (10mins) of video material for each participant. This meant that a further subset of words had to be chosen. As the goal was to probe visual-linguistic integration, a final set of 100 video clips was pre-selected  in such a way as to ensure the generation of a stimulus set containing both videoclips in which the visual information was highly relevant for processing and videoclips in which it was not. To this end, each frame was captioned in each video scene with the CLIP model and the text-based transformer BERT\cite{devlin_bert_2019} was used to calculate the semantic similarity between these captions and the word of interest. The maximum similarity score derived from this comparison was used for each scene as a measure of how relevant the preceding scene was for the upcoming word. Then, from each movie, 25 scenes were chosen from the low end of the distribution of these scores and 25 scenes from the high end of the distribution of these scores - resulting in 100 movie scenes in total, 50 from each movie. These were validated against the later collected human scores, showing that around 75\% of categorizations into ‘high’ or ‘low’ relevance matched with human judgements. The final set of stimuli and the code for determining them, can be found at https://github.com/ViktorKewenig upon publication. 

After the main experiment was conducted, 15 frames were extracted from each of the 25 video clips that received the highest predictability ratings by human participants for our follow up study. 

Both the main study and the follow up studies were made with the Gorilla task builder,\cite{anwyl-irvine_gorilla_2020} and the structure of both studies was as follows: after recruitment through Prolific, participants were directed to the study website. Participants were briefed on the aims of the study and asked for consent. Following the collection of demographic information, participants were instructed to complete the experiment in an environment with minimal distractions and with their phones turned off, wearing headphones. For the main study, before starting the experimental task, participants were run through the native eye tracking calibration for web cameras on Gorilla. The eye tracking software was calibrated three times in total – before the practice trials, after the practice trials, and during the halfway break – to ensure more reliable measurement even with a shift of head position during the task. Both experimental tasks consisted of 3 practice trials and 100 experimental trials. 

During the main study, in each trial, participants were first shown the word of interest corresponding to the upcoming scene and asked to pay careful attention to the upcoming video clip and think about how relevant that clip would be for predicting the word. After pressing a key to continue, participants were then shown the 6s movie scene. The video was shown in a window at the centre of the screen at a resolution of 720x1280 pixels (25 frames per second). Participants then used a 100-point slider (‘Low Relevance’ to ‘High Relevance’) to indicate how relevant the video clip was to the previously presented target word. This was repeated for all 100 trial scenes and took participants an average of around 30 minutes to complete. A self-paced resting break was provided halfway through the trials. After completing the experimental trials, participants were debriefed and paid for their time.

Following the main experiment, we conducted a detailed semantic annotation study of all segments (n=1500) that showed significant model-human attention alignment. Twenty-five independent raters were recruited to analyze each 400ms segment. For each segment, raters were shown the frame and the target word to be predicted, then asked to complete two tasks:

Category Selection: Raters identified which aspects were most informative for predicting the target word from a predefined set:
\begin{itemize}
\item Objects
\item Actions
\item People
\item Setting/Environment
\item Emotions Displayed
\item Target word's referent presence
\item None of the above
\end{itemize}
Description Task: For each selected category, raters provided concise (maximum three words) descriptions of the specific informative elements.

For example, if the target word was ``departure'' a rater might select ``Actions'', ``Emotions Displayed'', and ``Setting/Environment'', describing them as ``packing suitcase'', ``anxious facial expression'', and ``airport'' respectively.

Inter-rater reliability was assessed using Fleiss' $\kappa$ for each segment's category selections. The mean $\kappa$ across all segments was 0.82 (SD = 0.14, range = 0.53-0.94). For subsequent analyses, we used only segments showing high inter-rater reliability ($\kappa$ > 0.80, N = 873 segments). This threshold was chosen as it indicates "almost perfect" agreement in standard interpretations of $\kappa$ values \cite{landis1977measurement}. Reliability varied systematically across categories: highest for concrete features like "Objects" ($\kappa$ = 0.88) and "People" ($\kappa$ = 0.86), and lower for more abstract categories like "Emotions Displayed" ($\kappa$ = 0.71). Description task consistency was evaluated separately, with agreement considered valid when at least 80\% of raters used semantically equivalent terms (as determined by BERT similarity scores > 0.85).

For final analyses of semantic features driving attention alignment, we included only segments that met both the category selection reliability threshold ($\kappa$ > 0.80) and the description consistency criterion. This stringent filtering ensured that our conclusions about semantic drivers of attention were based on highly reliable annotations.

Human estimates of the target word's predictability were collected using an interactive slider. This measure of likelihood is intended to reflect the participant's intuitive understanding of the relationship between antecedent linguistic context, the visual cues in a scene and the subsequent semantic content. As we expected these intuitive judgements to vary between participants, outliers (any rating above or below 2 standard deviations) were not excluded from the analysis. A human ceiling score was calculated by correlating predictability estimates from each human participant against the $N-1$ (199) other participants and then averaging over these correlation values for all participants. This ceiling score allowed us to compare model scores against a baseline and they also functioned as an estimate of the reliability of individual humans. 

The gathering of human eye tracking data was done via the Gorilla eye tracking tool (version 2.0). This WebGazer-based tool achieved calibration accuracies within 100-150 pixels (approximately 2° of visual angle) and mean validation errors of less than 1° of visual angle across participants. The collected data, sampled at a rate of 60 Hz, was individually stored in an Excel file for each video-clip and each participant. To ensure reliability, calibration was performed three times during the experiment: before practice trials, after practice trials, and during the halfway break.

To isolate the contribution of visual and auditory information to predictability judgments (and eyegaze), we conducted an additional follow-up study with 100 participants. The procedure matched the main experiment, but all video clips were presented without audio. Participants viewed the same 6-second clips and were asked to rate the relevance of the visual information alone for predicting the target word on the same 0-100 scale. This allowed us to establish a baseline for purely visual predictability and compare it to the full audiovisual condition. Eye tracking data was collected using the same methodology as the main study. Importantly, this visual-only condition resulted in a lower human ceiling for eyetracking-attention overlap (mean inter-human correlation $\rho$ = 0.21 compared to $\rho$ = 0.32 in the audiovisual condition), indicating greater variability in how participants interpreted purely visual cues. This difference provided a baseline for evaluating how models integrate unimodal versus multimodal information.

 In the unimodal paradigm, LLaMA encodes and processes the textual input. A softmax over the next-word logits was computed for all labels in the movie. Logits are essentially the raw predictions that the model generates for the upcoming word, before they are converted into a more interpretable form, such as probabilities, through the softmax function. These logits were obtained by slightly altering the model's `forward' method and indexing into the final word logits. In the multimodal paradigm, the textual input was processed by the frozen LLaMA model, whereas the visual (frame-by-frame) information was sent to the adapter layer. LLaMA adapter uses a CLIP-based encoding for the visual information and then projects this encoding into an embedding space that can be processed by LLaMA. Again, predictability scores were extracted by computing a softmax over the next-word's logits for all labels in the movie, obtained from the model's `forward' method. A potentially confounding factor is that the adapter layer is itself fine-tuned—while the original LLaMA weights remain frozen—on a large instruction-following corpus that pairs CLIP image embeddings with chat-style prompts and textual answers, so part of the multimodal gain may reflect this additional instruction-tuning rather than visual grounding alone.

FLAVA-full \citep{singh2022flava} was expressly designed to operate in three modalities—text–only, image–only, and image,+,text—by sharing parameters across a text encoder (RoBERTa-base), a vision encoder (ViT-B/16) and a multimodal transformer that fuses both streams.

In the unimodal condition we supplied only the subtitle prompt. FLAVA skips the vision encoder in this setting and routes the CLS token together with the text sequence through the multimodal transformer, where a masked-language-model (MLM) head produces token-level logits. We masked each candidate label in turn, indexed its position in the logits tensor, and applied a softmax to obtain the probability that the label is the next word in the movie dialogue—exactly matching the procedure used for LLaMA.

In the multimodal condition we passed the same prompt together with the corresponding video frame. The frame is first tokenised into 14,$\times$,14 visual patches; these are embedded by the vision encoder and then concatenated with the text tokens before entering the multimodal transformer. Because the MLM head remains the same, the next-word logits are now conditioned on both linguistic and visual evidence, allowing us to extract predictability scores for every label via the identical soft-max step.

FLAVA is an ideal test-bed for comparing unimodal and multimodal prediction: a single set of decoder weights produces the logits in both settings, so any difference in predictability can be attributed entirely to the additional visual context rather than to architectural or training-objective changes.

To obtain GPT-4 predictability scores in the multimodal paradigm, OpenAI's API was used. The human instructions were used as prompt-based input to the API, together with the textual input. Temperature was set to 0, so as to obtain a deterministic outcome. In order to obtain multimodal predictability scores, the API was also used, except this time with the “GPT-4-V” (the visual) version of GPT-4. This allowed the uploading of the video scene as a as a `Graphics interchange format' (GIF) (which is a more dynamic representation of the video information compared to singular, frame-by-frame screenshots), together with the human instructions and the textual input (which human participants would listen to as the dialogue). Again, the temperature was set to 0 in order to obtain a deterministic outcome. 

Because GPT-4-V shares the same underlying parameters and decoding stack as the text-only GPT-4, any difference in predictability can be attributed solely to the added visual context.

The calculation of CLIP predictability scores for target words was carried out through a combination of image and text feature comparison with a custom Python script making use of the packages ‘PyTorch’\cite{paszke_pytorch_2019} and the Hugging Face ‘transformer’ library.\(^{48}\) Each frame of each movie clip was individually preprocessed by the standard, pre-trained CLIP preprocessor, transforming the visual data into a digestible form for the CLIP model's image encoder. This encoder generates a corresponding image feature vector, which provides a compact representation of the visual content in a form that can be easily compared with textual information. We pair each image feature vector with an array of text feature vectors. These text features vectors are derived from both the dialogue of each scene, represented as text prompts, and the collective word set present in the movie. The CLIP model tokenizes and encodes these text inputs, converting them into a format that mirrors the structure of the image feature vectors. For the purpose of our study we used both the the ‘clip-vit-base-patch32’ version and the `clip-RN50' versions of the model.\cite{noauthor_openaiclip-vit-base-patch32_nodate}
 
The calculation of the `predictability score' was carried out by comparing  these two feature sets, visual and linguistic. By computing the dot product of the image feature vector and each text feature vector, a raw similarity measure was derived. This raw similarity measure was then passed through a softmax function, refining it into a more interpretable similarity score. This score quantified the likelihood of the upcoming word, given all words in the movie, by matching each word to the visual content of each associated frame (or textual input in the multimodal paradigm). 

 Images were encoded through BLIP's ViT-based visual encoder (after resizing to 384x384 and applying standard normalization), while text was processed through the text encoder. The text-decoder then integrated these modalities using cross-attention. We extracted predictability scores by applying a softmax to the decoder's output logits, specifically indexing into the final position's logits where the model predicts the next word. For words requiring multiple tokens, we computed the geometric mean of the token probabilities to obtain a single word-level predictability score. All computations used the modest-sized "Salesforce/blip-image-captioning-base" pretrained model with cross-attention enabled in both visual and textual components, processing images frame-by-frame to maintain temporal consistency with other models.

There are both commonalities and differences between our model-based predictability measures and the human predictability scores we collected. A noteworthy difference lies in the approach to estimating predictability between LLaMA, FLAVA, and human participants. While LLaMA, FLAVA, and BLIP compute the likelihood of the next word directly based on textual or multimodal input, our human participants were asked to assess the \emph{relevance} of the visual-linguistic information for predicting an upcoming word, rather than directly predicting the next word. This process differs from the models' computations, as it does not require participants to generate specific next-word predictions. FLAVA is particularly informative here because the very same architecture can operate in text-only and image+text modes; any improvement when vision is added therefore cleanly isolates the contribution of visual context without confounding architectural changes. The reason we chose relevance ratings for the human experiment is both practical and theoretical. Predicting the next word from a six-second movie scene is inherently difficult for human participants and might not have yielded usable results. In our pilot studies, we found that participants' predicted words often did not match the actual next word or the model's top predictions. This mismatch made it challenging to correlate human predictions directly with model outputs, especially since human participants typically provide a single word without an explicit probability estimate, whereas language models generate a probability distribution over a range of possible next words.

Moreover, asking participants to generate multiple word predictions and estimate probabilities for each would significantly increase cognitive load and the duration of the experiment, potentially leading to participant fatigue and reduced data quality. To maintain within-subject consistency and collect data across many items, we opted for a more streamlined approach. Relevance ratings, on the other hand, allow participants to assess how well the provided context supports a specific upcoming word (the target word). This method effectively provides an estimate of the perceived probability of that word occurring next, enabling a direct comparison with the model's probability estimates for the same word. Relevance ratings capture a broader assessment of how well the context aligns with the potential upcoming word, accommodating the inherent uncertainty and variability in natural language processing.

From a theoretical perspective, relevance ratings reflect the cognitive flexibility humans exhibit when processing language, especially in complex, multimodal environments where exact predictions are difficult. While direct predictability tasks focus on word-level anticipation, requiring participants to pinpoint the exact next word and assuming a fine-grained understanding of the context, relevance judgments tap into probabilistic reasoning, acknowledging multiple plausible continuations. To alleviate concerns connected to these differences, we also included a prompt-based measure with GPT-4 using the same instructions provided to human participants. 

Regarding the CLIP model, it generated quantitative estimates of the likelihood of each word by comparing the feature vectors of the visual content in a frame with feature vectors of the associated linguistic context. The predictability measures extracted from CLIP and human participants therefore rely on a similar foundational principle: considering the similarities within the presented visual-linguistic information and understanding how these similarities influence the likelihood of the upcoming target word. This integration is an inherent part of human language comprehension, combining visual stimuli with linguistic expectations based on past experiences and world knowledge.

To directly test the causal role of attention mechanisms in human-model alignment, we conducted ablation experiments using CLIP ViT-B/32. We implemented two attention manipulations that preserved the model's architectural structure while disrupting the learned attention patterns.
\paragraph{Attention Manipulations}
We modified the Vision Transformer's attention mechanism by intercepting attention weights at each transformer block using forward hooks. Two ablation conditions were implemented:
\textbf{Uniform Attention:} All spatial attention weights were replaced with uniform values, such that each image patch received equal attention weight ($w_{ij} = \frac{1}{N}$ where $N$ is the number of patches). This manipulation removes spatial selectivity while preserving the attention mechanism's computational structure.
\textbf{Patch-Shuffled Attention:} Learned attention weights were randomly redistributed across spatial locations using a fixed random seed (seed=0). This manipulation preserves the magnitude and distribution of attention weights while disrupting their spatial correspondence to visual content.
Both manipulations were applied consistently across all 12 transformer layers. The model's vision encoder processed the same 100 videos used in the main analysis, generating probability distributions over the same 100 labels through contrastive similarity scoring with text embeddings.
\paragraph{Statistical Analysis}
We compared each ablation condition against the original learned attention using mixed-effects regression models. The dependent variable was human relevance ratings (HumanProb), with standardized model probability scores (ModelProb\_z) and attention condition as fixed effects. Random intercepts and slopes were included for participants to account for individual differences in model-human alignment patterns. The key test was the interaction between model confidence and attention condition, which assessed whether attention manipulations altered the relationship between model predictions and human judgments.
Model fitting used maximum likelihood estimation with the \texttt{statsmodels} package in Python. Significance was assessed at $\alpha = 0.05$, with effect sizes reported as standardized regression coefficients.

To explore the qualitative role of attention in the alignment between models and humans in the multimodal paradigm, we compared three types of attention matrices: BLIP's purely visual attention (from vision-only processing), CLIP's visual attention (biased through pre-training), and BLIP's cross-attention (integrating visual-linguistic information). All models use 12 transformer layers with multiple attention heads (16 for CLIP and BLIP's visual attention, 12 for BLIP's cross-attention), yielding multiple attention maps per input. For our analysis, the attention weights of all layers were utilised. All attention maps were reshaped to match the resolution of the human eye tracking heatmaps (the 1280 x 720 pixelrate of both movies).

All models operate on the principle of dividing the input image into a grid of patches. Each patch represents a segmented portion of the image, with the model's attention heads focusing on these discrete units rather than individual pixels. This patch-based approach allows the model to efficiently process and interpret the visual information by focusing on salient image segments. The models do not analyse the full-sized image at the original pixel resolution of 1280x720. Instead, CLIP downsamples images to 224x224 pixels and BLIP to 384x384 pixels, which are then divided into patches. For BLIP's cross-attention, we extracted weights where text features attend to image patches (in the text decoder layers, a 24x24 grid), averaging these weights across text tokens before upsampling.

A two-dimensional 'attention heatmap' was then generated for each attention head by reshaping the patch-scale attention weights back into a square matrix of the original image size (1280x720). Each cell in this matrix represented a region of the input, and its value indicates the amount of attention the model paid to that region. For BLIP's cross-attention maps, we used Lanczos interpolation to upsample the attention weights while maintaining the integrity of the attention patterns. The attention heatmaps were then averaged across all attention heads to create a single attention heatmap for each movie clip, similar to the averaged heatmaps generated from the human data.

In these 2-D matrices, each entry denotes the significance of a specific pixel in the visual input when generating a predictability score on the model side, and the duration a participant spent observing that specific pixel on the human side. To facilitate our analysis, we averaged attention matrices from model and human participants across every ten frames, yielding 15 averaged attention matrices per video for each participant and each model. Another reason for choosing to average the 6s video-clip into 15 segments (400ms each) is a fundamental difference between human and model processing. Humans do not instantly react to a stimulus; rather previous research on the visual world paradigm has estimated the saccade response to take between 100–250ms depending on the richness of the stimulus. \cite{dahan_chapter_2007, huettig_word_2005} As human gaze will therefore be lagged around that time for reaction to a new object or scene (while the model's attention patterns are near instantaneous), an average over 400ms was considered enough to smooth out that difference. Furthermore, 400ms is a relevant time-window for studies on predictive language processing, as the N400 EEG signal related to surprisal (or prediction error) can be measured 400ms after word-onset. \cite{kutas_thirty_2011} 

Another difference between attention in humans and visual transformers, is that visual transformer attention is widely distributed across the visual input, while human attention is relatively sparse (eye tracking data yields a single predicted pixel every 10ms). For this reason, we thresholded the model attention at a conservative 15th percentile of top values, and applied Gaussian smoothing to both model and human heatmaps. The sigma value in the Gaussian smoothing process determines the extent of smoothing and was chosen as that which maximises the difference between the human attention distribution and a null distribution. The null distribution was generated by randomly redistributing the attention values across human heatmaps. We chose the sigma value in this way so as not to bias later comparisons between human and model attention patterns. 

A probability distribution over the averaged and smoothed heatmaps was calculated by dividing each value in each heatmap by the total sum of values. Alignment between human eye tracking and model attention was quantified as the Spearman correlation between these probability distributions (15 for each of the 100 video clips). Since the Spearman correlation is based on ranks rather than actual values, it is less sensitive to outliers compared to the Pearson correlation. This is particularly useful for attention heatmaps, where there may be regions with exceptionally high or low attention that could skew the results of parametric correlations. A human ceiling correlation value for each of the 15 heatmaps per video clip was determined by correlating each participant's probability distribution with the $N-1$ (199) other probability distributions for this heatmap and taking the average of these correlations for each of the 15 segments per video clip.

To understand whether humans and the model focussed on different areas in the input when visual information was not relevant, we only considered those video clips in which the model-human correlations were negative. We defined a rectangle (640 x 360, half of all pixels) around the central pixel of the probabiltiy distributions of these segments as the central region and categorised the rest of the pixels as periphery. Next, we took the difference between the central and periphery areas for all probability distributions. A positive difference indicated that attention was centrally biased, whereas a negative difference indicated that attention was biased towards the periphery of the probability distribution. A t-test between all difference values on the model side and all difference values on the human side indeed suggested that CLIP tended to focus on periphery areas of the visual input  (t = -12.7, p \textless 0.001), whereas humans and BLIP with cross attention tended to focus on central areas (t = 9.9, p \textless 0.001).

\subsection*{Uncertainty-reduction (entropy) analysis}
We quantify how visual input constrains a model’s latent predictive state via token-level predictive entropy (bits). For a sentence \(s=(x_{1},\dots,x_{T})\), the text-only baseline is the mean entropy
\begin{equation}
H_{\text{text}}(s)
=
\frac{1}{T}\sum_{t} H\!\left(p(x_{t}\mid \text{text})\right).
\tag{3}
\end{equation}

For masked-LM architectures (FLAVA) we mask each non-special token and average the entropy of the masked-token distribution; for causal LMs (LLaMA) we average the next-token entropy \(H\!\left(p(x_{t}\mid x_{<t})\right)\); for BLIP we follow the same decoder-side protocol using the text decoder as the text-only baseline. In the multimodal (MM) condition we condition each architecture on every image \(i\) attached to \(s\) (15 per item), yielding \(H_{\text{mm}}(s,i)\). To summarise the strongest achievable visual constraint per item we define the \emph{best-pair entropy}
\begin{equation}
H_{\text{mm}}^{\star}(s)
=
\min_{i} H_{\text{mm}}(s,i)
\tag{4}
\end{equation}
and the \emph{entropy reduction}
\begin{equation}
\Delta H_{\text{model}}(s)
=
H_{\text{text}}(s) - H_{\text{mm}}^{\star}(s).
\tag{5}
\end{equation}

In parallel we quantify visual-set informativeness with CLIP by forming
\begin{equation}
p(i\mid s)
\propto
\exp\!\big(\tau \langle f_{\text{text}}(s), f_{\text{img}}(i) \rangle\big),
\tag{6}
\end{equation}
computing the image-set entropy
\begin{equation}
H_{\text{CLIP}}(s)
=
-\sum_i p(i\mid s)\log_{2} p(i\mid s),
\tag{7}
\end{equation}
and its deviation from uniform
\begin{equation}
\Delta H_{\text{CLIP}}
=
\log_2 15 - H_{\text{CLIP}}(s).
\tag{8}
\end{equation}

Entropies are micro-averaged across positions per sentence; 95\% CIs are estimated by nonparametric bootstrap (paired over sentences; 10{,}000 replicates).

\subsection*{Webcam-based Eye Tracking}

We recorded gaze with Gorilla’s WebGazer‑based system (v2.0) at 60Hz using commodity webcams. A 9‑point calibration/validation was run three times (at the beginning, in the middle, and at the end of the experiment). Across participants, calibration accuracy was typically 100–150px (~2° visual angle) with validation error < 1°. Participants were instructed to sit ~50–70cm from the display in a well‑lit room and minimize head movement. WebGazer uses a browser‑native, JavaScript implementation that maps pupil/eye‑region features to on‑screen coordinates via regularized linear regression and performs continual self‑calibration from user clicks and cursor trajectories; we adopt this model via Gorilla and analyze gaze at the region level rather than individual fixations to match the spatial granularity of transformer attention maps and the known precision limits of webcam tracking \cite{papoutsaki2016webgazer}.

To mitigate variability from lighting, camera quality, head pose, and camera‑to‑screen geometry—factors known to affect webcam ET accuracy—we applied objective QC: segments were discarded if valid samples fell below 70\%, if median post‑calibration error exceeded 2.5°, or if missingness/blink rate exceeded 30\% of the segment. Analyses used 400ms windows to accommodate human saccadic latency and align with the temporal grain of predictive language signals (e.g., the N400). Under comparable conditions, WebGazer reports errors of ~175–210px and an average 4.17° visual‑angle error relative to a (low‑cost) commercial tracker, which emphasises why coarse region‑level comparisons are appropriate here \cite{papoutsaki2016webgazer}.

More broadly, the webcam ET literature shows that performance improves with tailored models and calibration for the target device: on mobile phones/tablets, iTracker trained on the GazeCapture dataset (1,450 participants; ~2.5M frames) achieves 1.71–2.53cm error without calibration and 1.34–2.12cm with calibration, demonstrating that commodity cameras can reach centimeter‑level accuracy when models and data are matched to deployment \cite{krafka2016eyetracking}. We situate our study within this landscape: we use a standard WebGazer pipeline for large‑scale browser‑based testing, state limitations explicitly, and design analyses (region‑level alignment, temporal windows, QC thresholds) to be robust to environmental variability \cite{papoutsaki2015scalable}.

Finally, we note two recent developments that clarify the current position of the technology. First, webcam sensors are increasingly used not only for point‑of‑regard but also for pupillometry: PupilSense demonstrates practical webcam‑based pupil‑diameter estimation in everyday environments, though absolute measures remain sensitive to illumination and camera characteristics—considerations we address by instructing participants about lighting and by excluding high‑missingness segments \cite{shah2025pupilsense}. Second, in gaze‑target estimation from natural images, foundation‑model features now deliver state‑of‑the‑art results (e.g., Gaze‑LLE leverages a frozen DINOv2 encoder with a lightweight decoder), which highlights rapid gains in vision‑only gaze tasks; our work, by contrast, focuses on on‑screen gaze during reading/viewing with webcam ET, which has a different error profile but benefits from similar principles of robust calibration and aggregation at appropriate spatial scales \cite{ryan2025gazelle}.

\subsection*{Analyses}

All vectors were $z$-standardised across clips so that intercepts and slopes of the subsequent models could be interpreted in units of human standard deviations. Alignment and calibration were analysed with a linear mixed-effects model (\texttt{statsmodels}; random intercepts for \emph{Participant} and \emph{Video}, all two-way fixed terms for \emph{Model}, \emph{Modality} and \emph{Model\textsubscript{z}}). Predictive power was evaluated with leave-one-video-out cross-validation: for every block a simple linear regression fitted on the remaining 99 videos predicted the held-out clip, yielding cross-validated $R^{2}$ and RMSE. Item-level Pearson correlations were computed with \texttt{SciPy} \citep{virtanen_scipy_2020}. To express those correlations relative to behavioural reliability we used a leave-one-participant-out ceiling: each participant’s clipwise scores were correlated with the mean vector of the other 199 participants, Fisher-$z$ transformed, averaged, and back-transformed, giving a ceiling of $\rho_{\text{L1PO}}=0.58$. All ceiling-corrected correlations quoted in the paper are the raw Pearson $r$ divided by this value; the full set of mixed-model coefficients, cross-validated errors, and corrected correlations is provided in Supplementary Table S1.

Human- and model attention matrices were 1280x720 pixels, where each value at each pixel represents how much attention was paid to this pixel during each 400ms segment. Overlap between the models' attention weights and human eye tracking data was calculated by comparing the probability distributions over both model- and human heatmaps for each segment (1500 segments in total). Probability distributions were calculated by dividing each value in the heatmaps through the sum of all values. Each participants' probability distribution for each segment was correlated with the model's corresponding probability distribution using spearman rank correlation. A human ceiling was constructed by correlating each participant’s probability against that of the $N-1$ other participants and then averaging these correlations across participants (resulting in one human-ceiling correlation value per segment). Overall overlap was calculated by averaging all correlation values for human-model comparisons and for the human ceiling across segments. 

To probe where in the visual backbones category information emerges, we conducted a memory-efficient, layer-by-layer similarity analysis on three models: \textbf{CLIP ViT B/32} \citep{radford_learning_2021-1}, \textbf{BLIP base}, and the \textbf{BLIP} vision-only model \cite{li2022blip}. For each of the 1\,500 stimulus frames (1\,500 PNGs drawn from 15 temporal segments across \emph{Prestige} and \emph{The Usual Suspects}), we extracted one embedding \emph{per layer} and subsequently averaged patch tokens (CLS for BLIP) to obtain a fixed-length vector. Activations were collected with forward hooks and stacked into a tensor of shape $N \times L \times D$ (segments $\times$ layers $\times$ dimension).

For the human semantic reference, we used these percentage-based representations derived from the annotation study, where each segment was characterized by the proportion of participants who identified each of five categories (Objects, Actions, People, Setting/Environment, Emotions Displayed). Model semantic representations were extracted using multimodal embeddings: CLIP used concatenated image and text encodings (1024-dimensional), BLIP-Cross employed cross-modal embeddings from the full model architecture, and BLIP-Vision used vision-only representations (768-dimensional) from the visual backbone. To compute segment-level RSA contributions, we calculated for each segment $i$ the Spearman correlation $\rho_i$ between its similarity pattern to all other segments in human versus model representational space: $\rho_i = \text{corr}(\mathbf{h}_i, \mathbf{m}_i)$, where $\mathbf{h}_i$ and $\mathbf{m}_i$ are vectors of similarities between segment $i$ and all other segments for human and model representations, respectively. This approach quantified how much each individual segment contributes to the overall human-model representational alignment.

\subsection*{Semantic Understanding}
To evaluate the alignment between model attention maps and human fixation data, we computed three complementary saliency metrics following established protocols \cite{bylinskii_what_2019}. For each 400ms segment, we first normalized both the model attention weights (\(A_m\)) and human fixation density maps (\(A_h\)) to sum to one, creating probability distributions. The Similarity Score (SS) was computed using histogram intersection:
\begin{equation}
SS(A_m, A_h)
=
\sum_{i=1}^N \min(A_m^i, A_h^i),
\tag{9}
\end{equation}
where \(i\) indexes over all \(N\) pixels in the attention maps.

For the Normalized Scanpath Saliency (NSS), we standardized the model's attention map:
\begin{equation}
\hat{A}_m
=
\frac{A_m - \mu_{A_m}}{\sigma_{A_m}},
\tag{10}
\end{equation}
where \(\mu_{A_m}\) and \(\sigma_{A_m}\) are the mean and standard deviation of the model's attention values. NSS was then calculated as the mean attention weight at human fixation locations:
\begin{equation}
NSS
=
\frac{1}{n}\sum_{i=1}^n \hat{A}_m(x_i, y_i),
\tag{11}
\end{equation}
where \((x_i, y_i)\) are the coordinates of the \(n\) human fixations.

For Information Gain (IG), we first estimated a center bias baseline \(B\) using a Gaussian kernel density estimate (\(\sigma = 10\%\) of image width) over all fixation locations in our dataset. IG was then computed as:
\begin{equation}
IG
=
\frac{1}{n}\sum_{i=1}^n
\Big[
\log_2\!\big(\epsilon + A_m(x_i, y_i)\big)
-
\log_2\!\big(\epsilon + B(x_i, y_i)\big)
\Big],
\tag{12}
\end{equation}
where \(\epsilon = 10^{-20}\) prevents numerical instabilities.

For each metric, we computed bootstrap confidence intervals by resampling fixations within each segment 1000 times. Higher values indicate better alignment between model and human attention patterns across all metrics.

To investigate whether attention mechanisms that better align with human eye-tracking patterns also contribute more positively to semantic understanding, we conducted a segment-level analysis examining the relationship between attention-eyetracking overlap and RSA contributions. For each of the 1,500 stimulus-segment combinations, we correlated two key measures: (1) attention overlap scores representing the Spearman correlation between model attention weights and human fixation density maps, and (2) segment-level RSA contributions quantifying how much each segment contributes to the overall correlation between model and human semantic similarity matrices.

We analyzed the attention-semantic relationship with \textbf{temporal trend analysis} using rising vs. falling attention trajectories over time to understand whether attention dynamics affected semantic outcomes.
Statistical analysis employed Fisher-transformed correlations to assess the relationship between attention overlap scores and RSA contributions within each group, with significance testing at $\alpha = 0.05$. We used separate correlation analyses for each model and attention category to account for architectural differences.

\subsection*{Data and Code Availability}
All the data used in this study will be publicly available upon publication under: \href{http://osf.io/6whzq/?view_only=162085f95bab42b5a57b34b386143ba8}{CLIPPredict}. All code for data processing and analysis will be publicly available upon publication under the following GitHub repository: \href{http://github.com/ViktorKewenig}{github.com/ViktorKewenig}.

\subsection*{Contributions}
VK: conceived the study, secured funding for the online study, created stimuli and led the online experiment, wrote code for analyes, conducted data analyses, wrote and edited the paper.
AL: helped conceptualise analyses and edited the paper.
SAN: helped conceptualise analyses and edited the paper.
CE: helped lead the online study and edited the paper.
QLE: helped lead the online study and edited the paper.
RA: helped lead the online study and edited the paper.
GV: provided stipends for VK and CE, helped conceptualise the study, and edited the paper.
JIS: helped conceptualise the study and edited the paper.
All authors have read and approved the manuscript.

\subsection*{Competing Interests}
The authors declare no competing interests. 

\subsection*{Acknowledgments}
This work was supported in part by the European Research Council Advanced Grant (ECOLANG, 743035); Royal Society Wolfson Research Merit Award (WRM R3 170016) to GV; and Leverhulme award DS-2017-026 to VK and GV. Figures were created with BioRender.com

\bibliography{main}

\end{document}